%% file: main.tex
\documentclass[11pt]{article}
\usepackage[a4paper, margin=1.25in]{geometry}
\usepackage[T1]{fontenc}
\usepackage[normalem]{ulem}

\usepackage[ruled,vlined]{algorithm2e}
\usepackage{graphicx}
\usepackage{booktabs}
\usepackage{caption}
\usepackage{xcolor}
\usepackage{rotating}

\usepackage[dvipsnames,svgnames,x11names]{xcolor}

\colorlet{RED}{red}
\colorlet{WHITE}{white}
\colorlet{FORESTGREEN}{ForestGreen}
\colorlet{MAROON}{Maroon}

\usepackage[nocompress]{cite}
\usepackage{balance}
\usepackage{listings}

\lstdefinelanguage{XML}
{
basicstyle=\ttfamily\footnotesize,
  morestring=[b]",
  moredelim=[s][\bfseries\color{Maroon}]{<}{\ },
  moredelim=[s][\bfseries\color{Maroon}]{</}{>},
  moredelim=[l][\bfseries\color{Maroon}]{/>},
  moredelim=[l][\bfseries\color{Maroon}]{>},
  morecomment=[s]{<?}{?>},
  morecomment=[s]{<!--}{-->},
  commentstyle=\color{gray},
  stringstyle=\color{blue},
  identifierstyle=\color{red}
}

\usepackage{moreverb}
\usepackage[nounderscore]{syntax}
\graphicspath{{./figures/}}
\DeclareGraphicsExtensions{.pdf}
\usepackage[cmex10]{amsmath}
\usepackage{amssymb}
\usepackage{mathtools}
\usepackage{amsthm}
\usepackage{amsfonts}

\usepackage{subfig}

\usepackage{multirow}
\usepackage{rotating}
\usepackage{booktabs}
\usepackage{colortbl}
\usepackage{tablefootnote}

\usepackage{array}
\newcolumntype{L}[1]{>{\raggedright\let\newline\\\arraybackslash\hspace{0pt}}m{#1}}
\newcolumntype{C}[1]{>{\centering\let\newline\\\arraybackslash\hspace{0pt}}m{#1}}
\newcolumntype{R}[1]{>{\raggedleft\let\newline\\\arraybackslash\hspace{0pt}}m{#1}}

\usepackage[pdftex,colorlinks=true,urlcolor=blue,citecolor=blue]{hyperref}
\usepackage{xspace}

\usepackage{enumitem}

\usepackage[utf8]{inputenc}
\usepackage[T1]{fontenc}
\usepackage{url}
\usepackage{booktabs}
\usepackage{amsfonts}
\usepackage{nicefrac}
\usepackage{lmodern}
\usepackage{microtype}
\usepackage{lipsum}
\usepackage{rotating}
\usepackage{multirow}

\newcommand{\uvhw}{BMD-45\xspace}

\newcommand{\mapFiftyNinetyFive}{mAP@\allowbreak$0.50$:\allowbreak$0.95$\xspace}
\newcommand{\mapFifty}{mAP@\allowbreak$0.50$\xspace}
\newcommand{\mapSeventyFive}{mAP@\allowbreak$0.75$\xspace}

\begin{document}

\title{\uvhw: A Large-Scale CCTV Vehicle Detection Dataset for Urban Traffic in Developing Cities \footnote{Accepted at CVPR 2026 Findings Track. To appear in the IEEE/CVF Conference on Computer Vision and Pattern Recognition 2026
}}

\author{
Akash Sharma$^{1,\dagger}$, Chinmay Mhatre$^{4,\dagger}$, Sankalp Gawali$^{1}$, \\
Ruthvik Bokkasam$^{1}$, Brij Sharma$^{3}$, Vishwajeet Pattanaik$^{4}$, \\
Punit Rathore$^{2,4}$, Raghu Krishnapuram$^{3,4}$, Vijay Gopal Kovvali$^{4}$, \\
Anirban Chakraborty$^{1}$, Yogesh Simmhan$^{1,2}$ \\~\\
    \small $^{1}$Department of Computational and Data Sciences (CDS)\\
    \small $^{2}$Robert Bosch Center for Cyber Physical Systems (RBCCPS)\\
    \small $^{3}$Center of Data for Public Good (CDPG)\\
    \small $^{4}$Centre for Infrastructure, Sustainable Transportation \& Urban Planning (CiSTUP)\\
    \em Indian Institute of Science, Bengaluru, India\\~\\
    \texttt{Email: \{akashsharma, chinmayp1, simmhan\}@iisc.ac.in}\\
    \small $^{\dagger}$Equal contribution
}

\date{}
\maketitle
\begin{abstract}
Robust vehicle detection from fixed CCTV cameras is critical for Intelligent Transportation Systems. Yet existing benchmarks predominantly feature relatively homogeneous, highly organized traffic patterns captured from ego-centric driving perspectives or controlled aerial views. This regional and sensor view bias creates a significant gap. Models trained on datasets such as UA-DETRAC and COCO struggle to generalize to the dense, heterogeneous, disorganized traffic conditions observed in rapidly developing urban centers in emerging economies. To address this limitation, we introduce \uvhw, a large-scale dataset comprising 480K bounding boxes annotated over 45K images captured from over 3.6K operational Safe City CCTV cameras. \uvhw contains $14$ fine-grained vehicle categories, including region-specific modes such as auto-rickshaws and tempo travellers, which are not present in existing benchmarks. The dataset captures real-world deployment challenges, including extreme viewpoint variation, occlusion, and vehicle density. We establish comprehensive baselines using state-of-the-art detectors and reveal a striking domain gap: models fine-tuned on UA-DETRAC achieve only $33.6\%$ mAP@$0.50$:$0.95$, compared to $83.8\%$ when trained in-domain on \uvhw, representing a $2.5\times$ improvement that persists even when accounting for novel vehicle classes. This performance gap underscores the critical need for geographically diverse traffic benchmarks and establishes \uvhw as a baseline for developing robust perception systems in underrepresented urban environments worldwide.
The dataset is available at: \href{https://huggingface.co/datasets/iisc-aim/BMD-45}{https://huggingface.co/datasets/iisc-aim/BMD-45}.

\end{abstract}

\section{Introduction}
Vehicle detection benchmarks predominantly fall into three categories based on the capture platform:~\emph{(i) ego-centric datasets} from vehicle-mounted cameras that capture side and rear views;~\emph{(ii) aerial datasets} from drones that provide top-down perspectives; and~\emph{(iii) fixed-camera datasets} from urban safety and infrastructure cameras that offer a raised and wide-angle viewpoint. Each of these viewpoints presents distinct characteristics; for instance, ego-centric cameras capture nearby vehicles with rich detail, aerial platforms provide a bird's-eye coverage, while fixed urban safety cameras occupy a middle ground with oblique angles, moderate elevation, and high instance density.

Among these, fixed cameras are particularly useful for \textit{Intelligent Transportation Systems (ITS)} as cities already operate thousands of CCTV cameras for urban safety and traffic management~\cite{Tang2019_CityFlow, Peppa2021, Zhou2025} that provide persistent wide-area coverage for incident response, which can be leveraged for traffic flow estimation~\cite{Pi2022, Yu2023, Myagmar2023}. While recent deep learning methods achieve competitive performance on benchmark datasets~\cite{Geiger2012_KITTI, Tang2019_CityFlow, Yu2020_BDD100K}, they often fail on fixed-camera viewpoints due to domain shift — models trained on ego-centric dashcam footage struggle to deal with the distinct characteristics of fixed, elevated viewpoints where vehicles appear smaller, more numerous, and are frequently occluded~\cite{Chen2018, Zhou2025}.

This is due to existing datasets inadequately representing fixed-camera viewpoints. Despite their operational importance, fixed-camera datasets remain limited in scale, diversity, and geographic coverage~\cite{Huang2025}. Most existing corpora originate from regions with structured traffic patterns and homogeneous vehicle classes, lacking representation of complex traffic flows and vehicle morphologies, such as auto-rickshaws ("tuk tuk" or three-wheelers), Light Commercial Vehicles (LCVs), and minibusses, which are prevalent in the Global South~\cite{Varma2019_IDD}. This geographic bias, combined with modest dataset sizes and coarse taxonomies, limits the generalizability of vehicle detection models for global deployments~\cite{Liu2023}, especially in developing countries.

\begin{figure*}[t!]
  \centering
  \includegraphics[width=\linewidth]{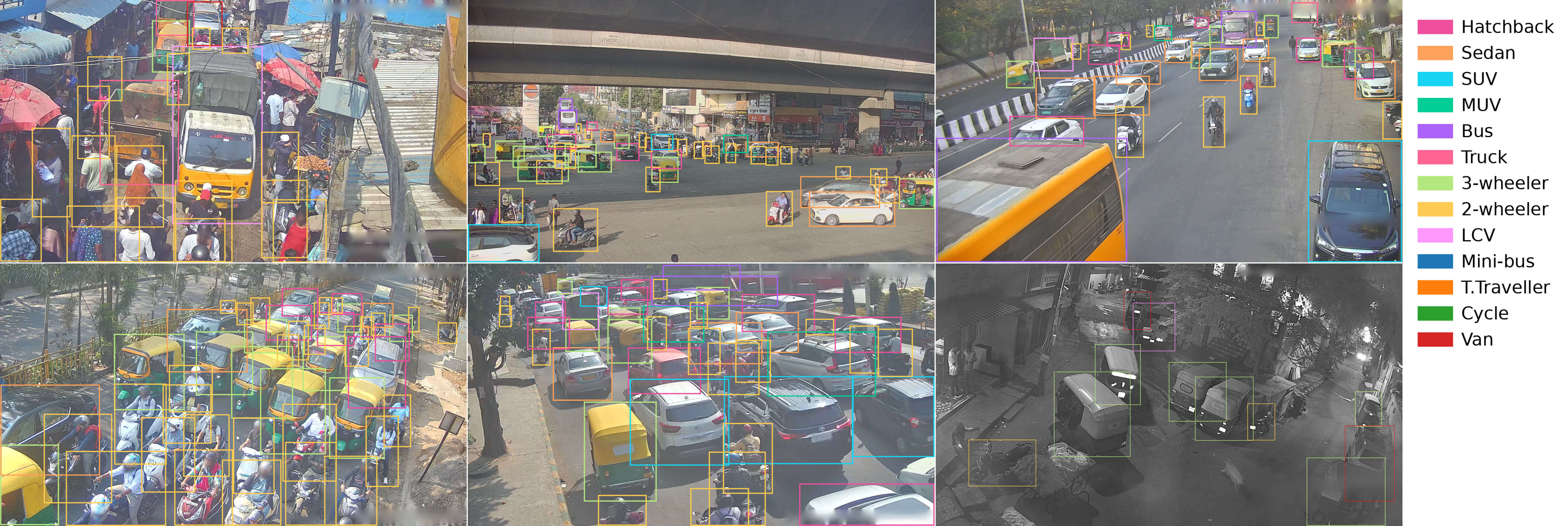}
   \caption{Annotated sample images from the \uvhw dataset illustrating the variety of vehicle classes and camera views.}
   \label{fig:sample_images}
\end{figure*}

To address these limitations, we introduce the Bengaluru Mobility Dataset (\uvhw), a large-scale vehicle detection dataset captured from thousands of operational CCTV cameras in Bengaluru, a megacity consistently ranked among the most traffic-congested in the world~\cite{TomTom2025}. The dataset comprises $\approx45,986$ images with 2 million crowdsourced bounding-box annotations spanning 14 fine-grained vehicle classes common in India, of which 480,000 unique annotations are obtained through majority voting. Representative examples are shown in Figure~\ref{fig:sample_images}.

Although the data originates from a single city, this geographic focus is consistent with established benchmarks such as Cityscapes~\cite{Cordts2015_Cityscapes} and CityFlow~\cite{Tang2019_CityFlow}, both of which are region-specific yet widely adopted. Moreover, prior works on Indian traffic datasets~\cite{ITD_COMSNETS24, JATAYU_CONECCT24} note that Indian vehicle classifications transfer well to South and Southeast Asian countries due to shared vehicle types (auto-rickshaws, electric-rickshaws, motorized two-wheelers) and centralized manufacturing standards.

Beyond geographic relevance, \uvhw also emphasizes visual diversity and annotation quality. Frames are sampled from long-duration CCTV recordings using spatial and temporal diversity criteria, and challenging scenes are further selected through difficulty scoring and model disagreement analysis. Annotation fidelity is maintained via a multi-stage crowdsourcing pipeline with redundant labeling and majority-vote aggregation. The resulting 14-class taxonomy captures vehicle-type distinctions relevant to regional traffic composition and ITS applications, including categories absent from existing benchmarks.

To validate the dataset's utility, we establish comprehensive baselines using state-of-the-art (SOTA) detectors, including YOLOv12~\cite{Tian2025_YOLOv12}, RT-DETRv2~\cite{Lv2024_RTDETRv2}, D-FINE~\cite{peng2025dfine}, RF-DETR~\cite{Robinson2025_RFDETR}, and Grounding DINO~\cite{Liu2024_DINO}. We evaluate these models using both validation and test splits of \uvhw, as well as cross-dataset generalization by training on \uvhw and evaluating on three SOTA benchmark datasets: UA-DETRAC~\cite{Wen2020_UADETRAC}, IDD~\cite{Varma2019_IDD}, and TrafficCAM~\cite{Deng2025_TrafficCAM}. For fair comparison across datasets with differing taxonomies, we define a mapping protocol that merges our fine-grained classes into coarse categories while preserving the original dataset definitions for ambiguous classes. Our findings demonstrate that models trained on \uvhw achieve robust performance on held-out test data and improve fixed-camera detection performance in cross-dataset evaluations, confirming the dataset's value for city-scale ITS deployment.

In summary, our work makes three primary contributions: \emph{(1)} A large-scale vehicle detection dataset from infrastructure cameras featuring 45K images, 480K annotations, and 14 fine-grained classes, with emphasis on underrepresented vehicle types and challenging CCTV conditions; \emph{(2)} Comprehensive benchmarks using modern detectors with rigorous evaluation protocols for cross-dataset generalization and taxonomy-aware class mapping; and, \emph{(3)} Demonstration that training on \uvhw improves fixed-camera detection performance across deployment contexts.

\section{Related Works}

\begin{table}[htbp]
\centering
\footnotesize
\caption{Comparison of SOTA vehicle detection and tracking datasets from fixed camera and CCTV viewpoints.}
\label{tab:vehicle_datasets}
\resizebox{\linewidth}{!}{
\begin{tabular}{@{}lllcccccc@{}}
\toprule
\textbf{Dataset} & \textbf{Venue (Year)} & \textbf{Task$^\dagger$} & \textbf{PoV$^\ddagger$} & \textbf{\#Frames} & \textbf{\#Annotations} & \textbf{\#Vehicle classes} & \textbf{\#Cameras} & \textbf{Location} \\
\midrule
\midrule
IDD~\cite{Varma2019_IDD} & WACV 2019 & D, S & E & 10k & $\ 111.3k$ & 9 & - & IN \\
UA-DETRAC~\cite{Lv2024_RTDETRv2} & CVIU 2020 & D, M & F & 140k & $1.21M$ & 4 & 24$^\ast$ & CN \\
TrafficCAM~\cite{Deng2025_TrafficCAM} & T-ITS 2025 & S & FC & 4.3k & $84.2k$ & 9 & NA & IN \\
\uvhw \textit{(Our)} & -- & D & FC & 45k & $481.9k$ & 14 & 3679 & IN \\
\bottomrule
\end{tabular}
}

\caption*{\footnotesize
$^\dagger$\textbf{Task Abbreviations:} D - Detection, S - Segmentation, M - Multi-object tracking. $^\ddagger$\textbf{Point of View Abbreviations:} E - Ego-centric; F - Fixed camera (non-CCTV); FC - Fixed CCTV cameras. $^\ast$\textbf{Camera count:} UA-DETRAC contains sequences recorded at 24 distinct locations, which we treat as separate fixed-camera viewpoints.
}
\end{table}

\subsection{Fixed Surveillance Datasets}
Fixed-camera datasets remain limited in scale, diversity, or accessibility. Among them, UA-DETRAC~\cite{Wen2020_UADETRAC} represents the current SOTA in scale, providing $\approx 140,000$ frames with $1.21$ million bounding boxes and $8,250$ unique vehicle identities from highway cameras in China. However, it was designed for multi-object tracking rather than detection, emphasizing temporal continuity over spatial diversity. This temporal redundancy limits the diversity of appearances needed for robust detection. Additionally, UA-DETRAC employs a coarse four-class taxonomy (car, bus, van, others) and focuses on structured highway traffic, limiting generalization to diverse urban scenarios and region-specific vehicle types.

The recent TrafficCAM~\cite{Deng2025_TrafficCAM}, on the other hand, addresses geographic diversity by capturing Indian CCTV footage with $4,402$ frames, including auto-rickshaws and region-specific vehicles, similar to our target deployment context. However, its small scale limits utility for training modern detectors. Furthermore, it focuses on semantic segmentation rather than instance detection, and leverages tracking-based annotation propagation, which may introduce noise in crowded scenes.

Other fixed-camera resources like CityFlow~\cite{Tang2019_CityFlow} use single-class ``vehicle'' annotations unsuitable for type classification, while CityCam~\cite{Zhang2017_CityCam} remains access-restricted. The fixed-camera landscape thus presents a trade-off: scale with low spatial diversity and coarse taxonomy (UA-DETRAC), or regional relevance and fine-grained classes but a lack of scale (TrafficCAM). 

\subsection{Cross-Domain Transfer from Ego-Centric Datasets}
The limited availability of fixed-camera datasets raises the question of whether models can be trained on alternative datasets and transferred to a fixed-camera context.
Ego-centric datasets like KITTI~\cite{Geiger2013_KITTI}, Cityscapes~\cite{Cordts2015_Cityscapes}, BDD100K~\cite{Yu2020_BDD100K}, nuScenes~\cite{Caesar2020_nuScenes}, and Waymo Open~\cite{Sun2020_Waymo} capture forward-facing views from vehicle-mounted cameras and offer significantly larger scale than existing fixed-camera resources. Among these, the Indian Driving Dataset (IDD)~\cite{Varma2019_IDD} is particularly relevant, as it was captured on Indian roads and includes $10,003$ semantically segmented images, including auto-rickshaws and unstructured traffic. However, ego-trained models transfer poorly to fixed (exocentric) cameras~\cite{Thatipelli2025} due to a fundamental viewpoint mismatch: a dashcam perspective ($1$--$2$~m elevation, horizontal angles) differ from a pole-mounted CCTV ($4$--$8$~m elevation, oblique angles~\cite{Klein2006}), altering vehicle aspect ratios, visible surfaces and occlusion patterns~\cite{Xiang2013}. We confirm this in our experiments (see~\ref{subsec:evaluation_existing_datasets}). Additionally, IDD focuses on semantic segmentation rather than bounding-box detection. These domain gaps limit the practical utility of transfer learning from ego-centric sources for fixed-camera deployments.

Recent surveys note this regional and modality skew in vehicle perception datasets, with limited fixed-camera coverage outside of developed regions~\cite{Liu2023}. For vehicle detection in unstructured urban traffic, the operational reality in Indian cities, with existing fixed-camera resources, inadequately balances scale, taxonomy granularity, and geographic diversity. \uvhw addresses these gaps by combining scale (45K images, 480K boxes), fine-grained taxonomy (14 classes), and diversity-driven sampling that prioritizes unique appearances across cameras and conditions, providing a purpose-built benchmark for CCTV-based ITS deployment.

\subsection{Comparative Evaluation with Existing Datasets}
\label{subsec:evaluation_existing_datasets}

\begin{figure}[t]
    \centering
    \includegraphics[width=1\columnwidth]{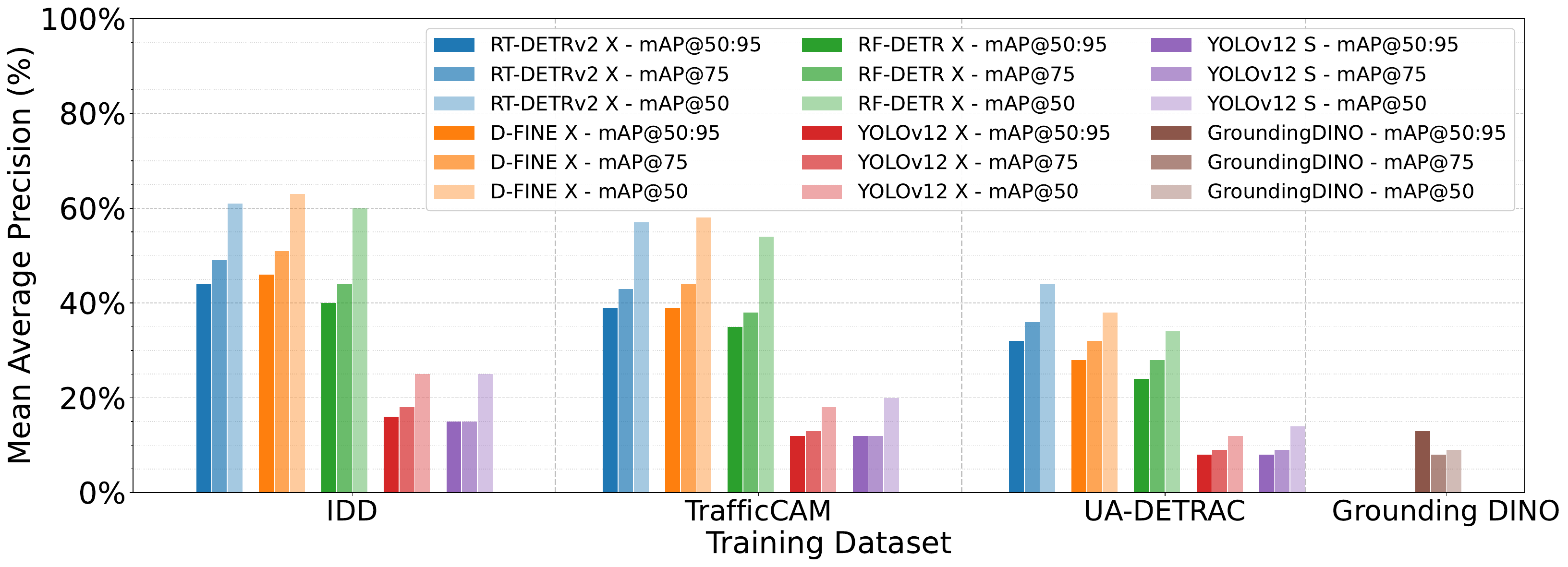}
    \caption{Cross-dataset generalization of object detectors on our expert-annotated reference set. Models trained on existing datasets exhibit substantial performance degradation when tested on the expert-annotated reference set.}
    \label{fig:cross_dataset_performance}
\end{figure}

To assess the adequacy of existing datasets for detecting and classifying vehicles in CCTV traffic images from India, we fine-tune SOTA object detectors on UA-DETRAC~\cite{Wen2020_UADETRAC}, TrafficCAM~\cite{Deng2025_TrafficCAM}, and IDD~\cite{Varma2019_IDD} using a common protocol described in Section~\ref{sec:comparison_protocol}. As outlined in Table~\ref{tab:vehicle_datasets}, the available datasets vary in scale, viewpoint, and annotation detail, and none offer both broad camera-view coverage and fine-grained vehicle classes. Each model was fine-tuned only on its respective dataset and then evaluated on an expert-annotated benchmark of $3,000$ CCTV images from Bengaluru with diverse camera views. As reported in Figure~\ref{fig:cross_dataset_performance}, the best IDD-trained model (D-FINE X) attains a \mapFiftyNinetyFive of $0.46$, while the best TrafficCAM-trained models (RT-DETRv2 X and D-FINE X) reach $0.39$. GroundingDINO, evaluated in a zero-shot setting, attains $0.13$, which is close to the weakest supervised baselines. These indicate that both supervised cross-dataset transfer and open-set zero-shot detection are far from satisfactory for our use case.

Models trained on UA-DETRAC, which contains structured highway scenes and a coarse four-class vehicle taxonomy, do not capture the richer set of vehicle types and dense urban layouts in our data. IDD-trained models transfer better in terms of \mapFiftyNinetyFive, but fall short of practical accuracy, which we attribute to the strong viewpoint gap between ego-centric dashcam footage and elevated CCTV cameras.

\begin{figure}[t]
    \centering
    \includegraphics[width=0.8\linewidth]{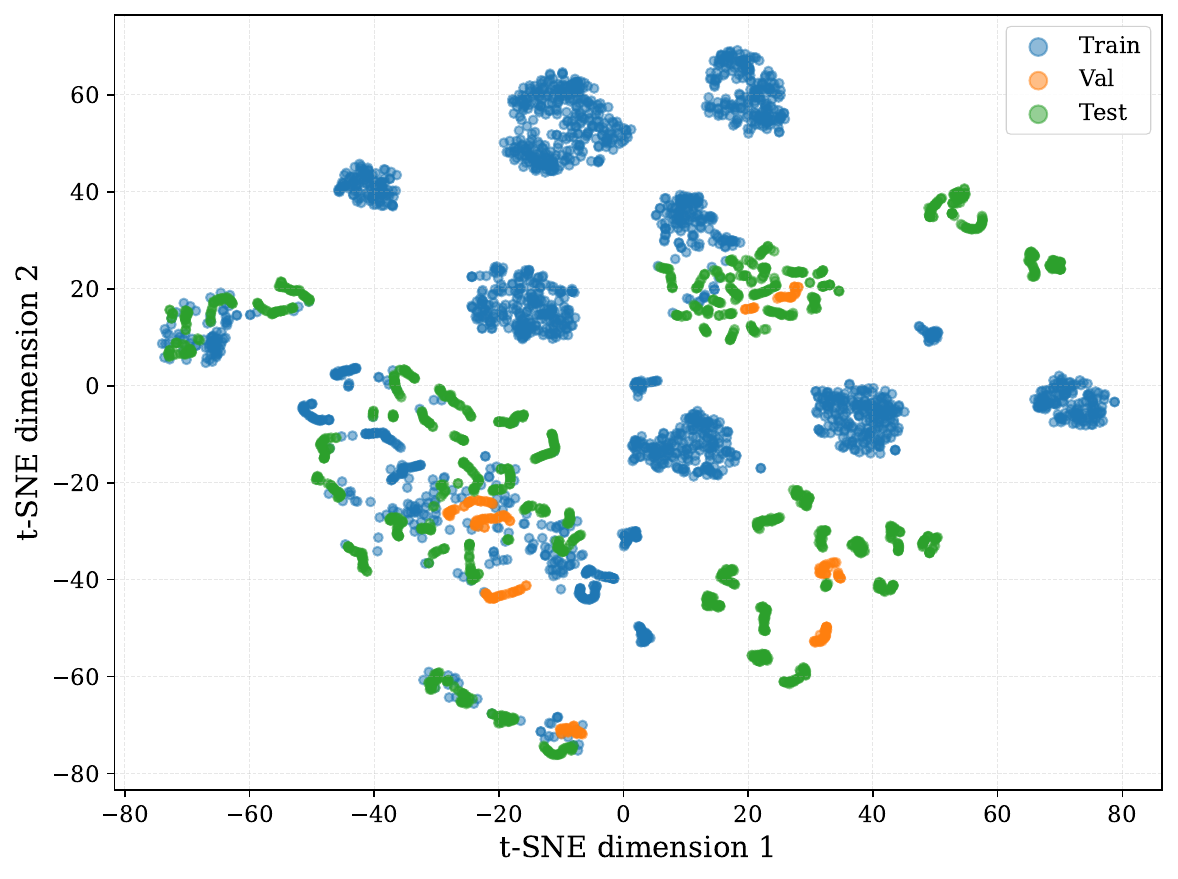}
    \caption{t-SNE visualization of DINOv3 frame embeddings for TrafficCAM; colored by split. }
    \label{fig:tsne-trafficcam-plot}
\end{figure}

TrafficCAM, despite sharing Indian geography and CCTV viewpoints, achieved lower \mapFiftyNinetyFive than the best IDD-trained model, despite IDD being ego-centric. We attribute this performance difference to TrafficCAM's limited scale and low viewpoint diversity; verified by the t-SNE projections of DINOv3 embeddings (Figure~\ref{fig:tsne-trafficcam-plot}), which reveal fewer than $10$ tight clusters with train/val/test samples mixed within the same clusters, indicating camera-view leakage across splits. A quantitative comparison of viewpoint diversity using Vendi Scores is provided in Section~\ref{sec:data_collection}.

\section{The \uvhw Dataset}

The proposed \uvhw dataset comprises $45,986$ high-resolution images ($1920\times1080$ RGB) with $481,947$ bounding box annotations across $14$ fine-grained vehicle classes. The images are sampled from $3,679$ cameras in Bengaluru's Safe City camera network, capturing the dense, heterogeneous, and disorganized traffic characteristic of Indian urban environments. Representative examples of these scenes and their annotations are shown in Figure~\ref{fig:sample_images}. The dataset is partitioned into \textit{training} ($35,792$ images, $375,003$ annotations), \textit{validation} ($10,194$ images, $106,944$ annotations), and \textit{test} ($5,110$ images, $92,065$ annotations) splits. To support rigorous benchmarking, the ground-truth annotations of test images will be withheld while the images themselves will be publicly released, similar to COCO~\cite{lin2014coco}.

\begin{figure}[t]
    \centering
    \includegraphics[width=0.8\linewidth]{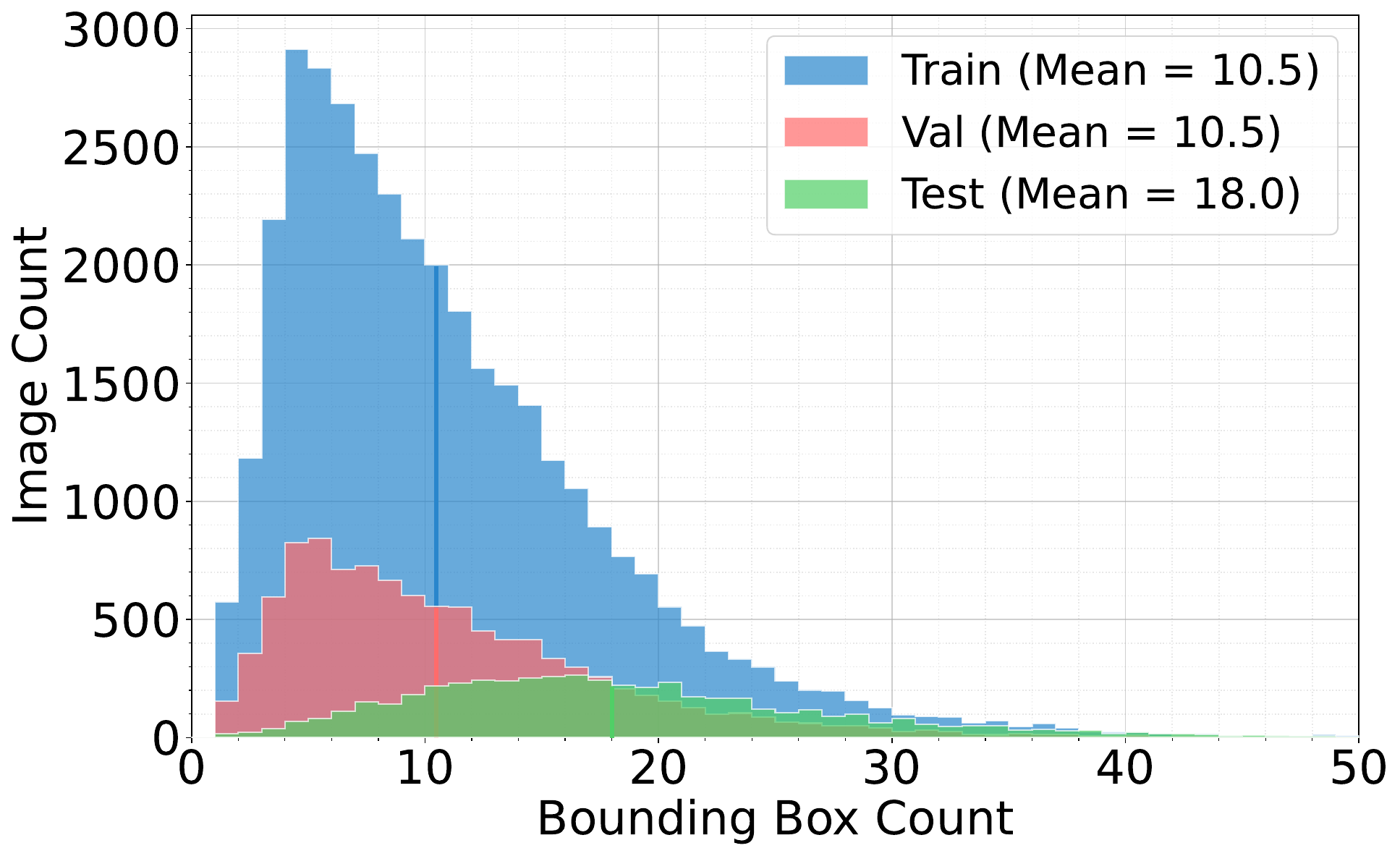}
    \caption{Distribution of \# of bounding boxes per image present in the dataset split (Train, Test, Val) in \uvhw.}
    \label{fig:bbox_count_per_image_aggregate}
\end{figure}

Figure~\ref{fig:bbox_count_per_image_aggregate} illustrates the distribution of annotations per image across splits. To meaningfully evaluate robustness under real-world complexity, we curate the test split from the most reliably annotated and most challenging subset of the dataset. We first identify the $10,000$ frames annotated by the largest number of independent annotators, ensuring high-confidence ground truth, and from these select the $5,110$ frames with the greatest image difficulty (explained in Section~\ref{sec:difficulty_sampling}). This prevents distribution skew: train and val maintain similar category distributions, while the test set is intentionally harder and contains denser, more complex scenes. This design yields a controlled yet demanding benchmark that exposes performance differences not captured by uniformly sampled data.

\subsection{Vehicle Class Design}
We define a 14-class hierarchical taxonomy that captures region-specific vehicle types in India while enabling aggregation to coarser categories. These are based on regulatory classifications and marketing/product segmentation.

\noindent\textbf{Vehicle Classes:} \emph{Bicycle}; \emph{Two-Wheelers} [Motorbike]; \emph{Three-Wheelers} [Auto-rickshaw]; Car [\emph{Hatchback}, \emph{Sedan}, \emph{Sport Utility Vehicle (SUV)}, \emph{Multi-Utility Vehicle (MUV)}]; Bus [\emph{Bus}, \emph{Mini-Bus}]; \emph{Van}; Commercial Vehicle [\emph{Light Commercial Vehicle} (LCV), \emph{Tempo-Traveller}]; \emph{Truck}; \emph{Other} [Unclassified].

\begin{figure}[t]
    \centering
    \includegraphics[width=0.8\linewidth]{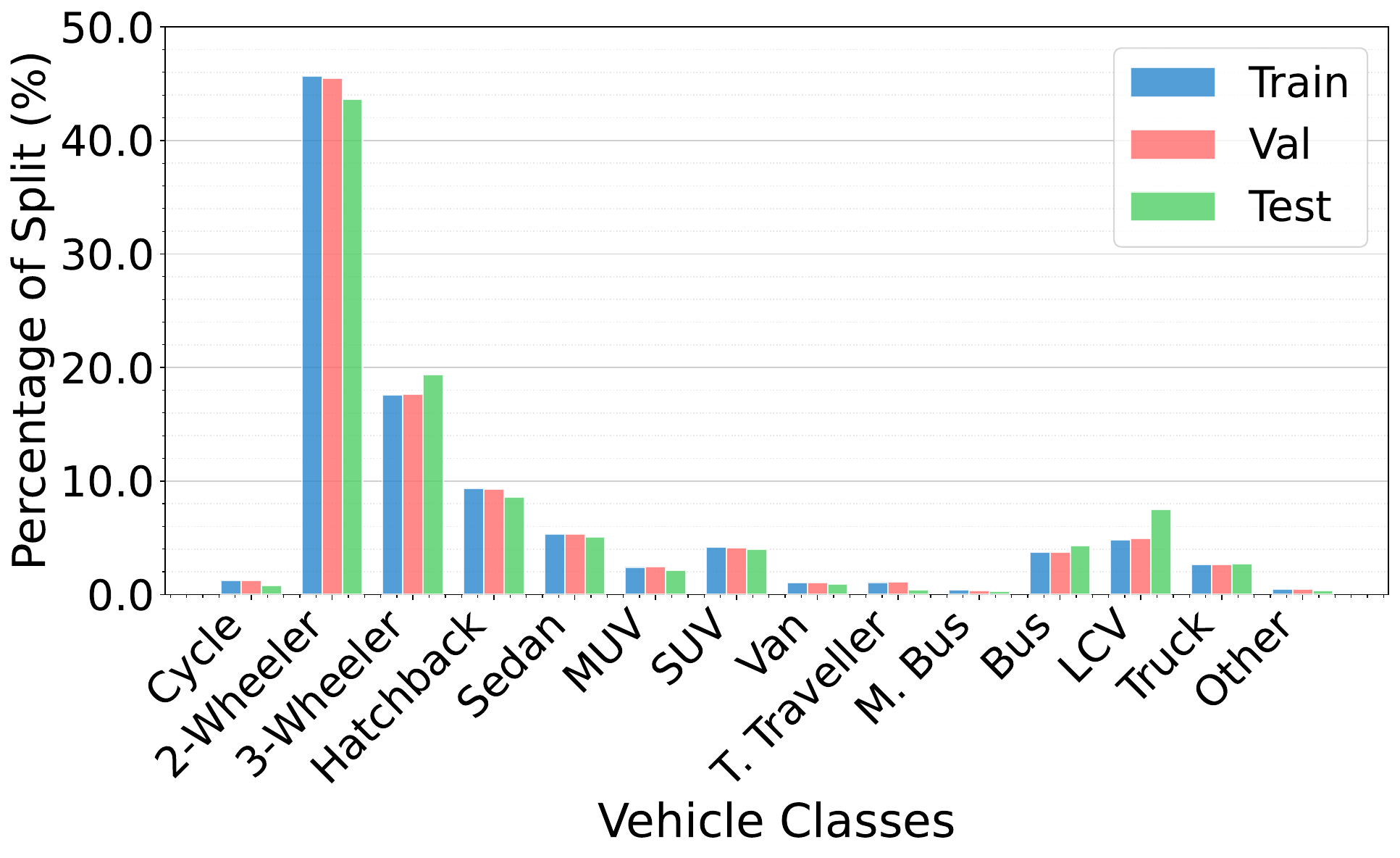}
    \caption{Distribution of \# of bounding boxes per class present in the dataset split (Train, Test, and Val) in \uvhw.}
    \label{fig:instances_per_class_aggregate_linear}
    
\end{figure}

This fine-to-coarse design aims to support flexible evaluation, allowing models trained on granular categories to be aggregated to parent classes (e.g., all car subtypes → ``Car''), enabling fair and uniform comparison with coarser taxonomies. Further, classes unique to the region, such as ``tempo-traveller'' and ``auto-rickshaw'', distinguish this dataset from Western-centric benchmarks. Figure~\ref{fig:instances_per_class_aggregate_linear} shows the class distribution across splits.

\subsection{Data Collection and Sampling Strategy}\label{sec:data_collection}

\begin{figure}
  \centering
  \includegraphics[width=\linewidth]{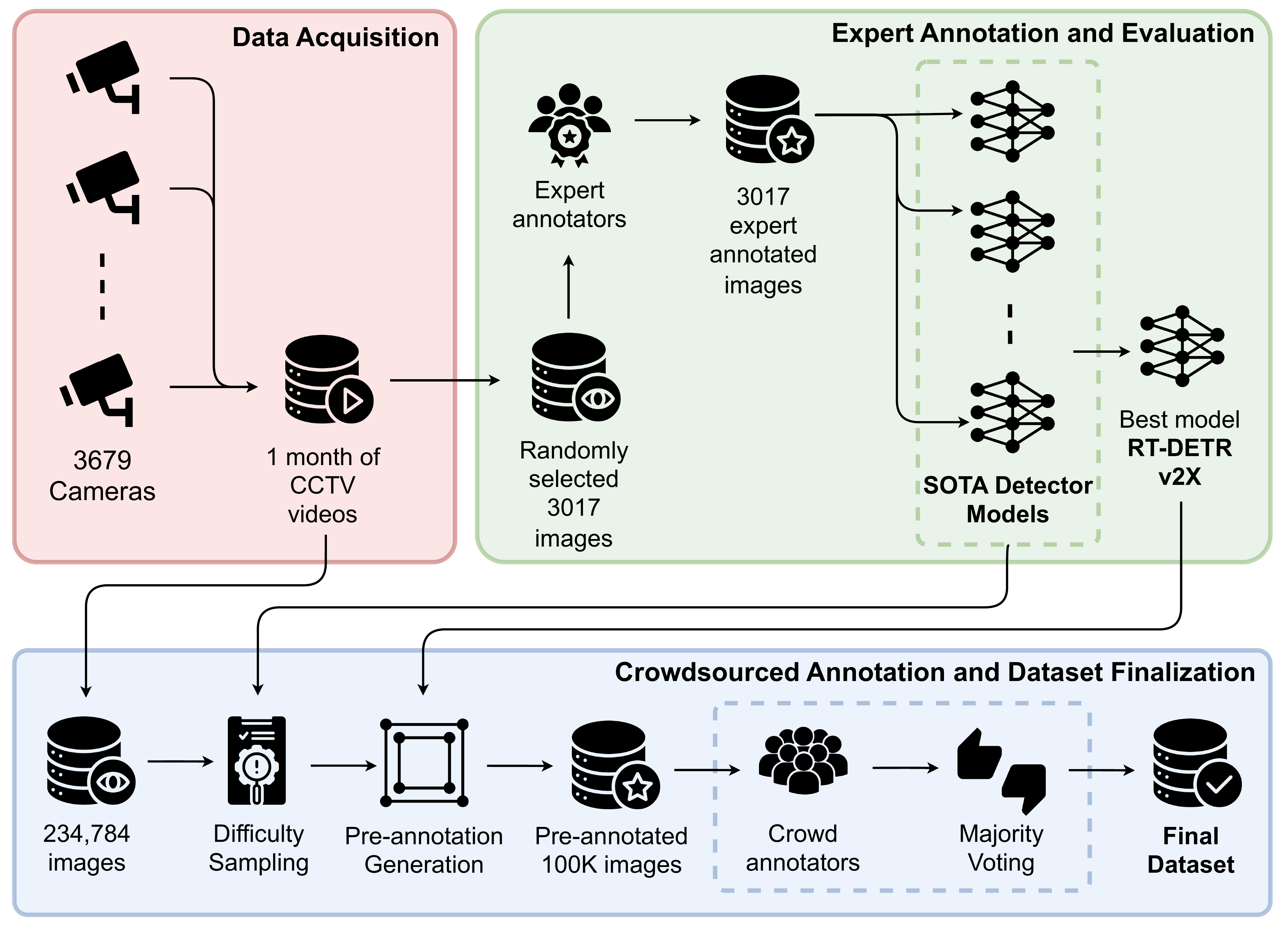}
  \caption{The \uvhw dataset collection and annotation pipeline.}
  \label{fig:crowdsource_workflow}
\end{figure}

\subsubsection{Camera Infrastructure and Temporal Coverage}
Figure~\ref{fig:crowdsource_workflow} shows the overall workflow of the crowdsourced annotation pipeline. We collected continuous CCTV footage from $3,679$ cameras distributed across Bengaluru's road network covering $\approx 740~km^2$ over a 28-day period spanning January to February 2025. These cameras monitor major intersections, arterial roads, and highway segments, providing comprehensive spatial coverage. However, their elevated mounting, dictated by public safety requirements, introduces significant detection challenges, as discussed previously.

\begin{figure}[t]
    \centering
    \includegraphics[width=0.8\linewidth]{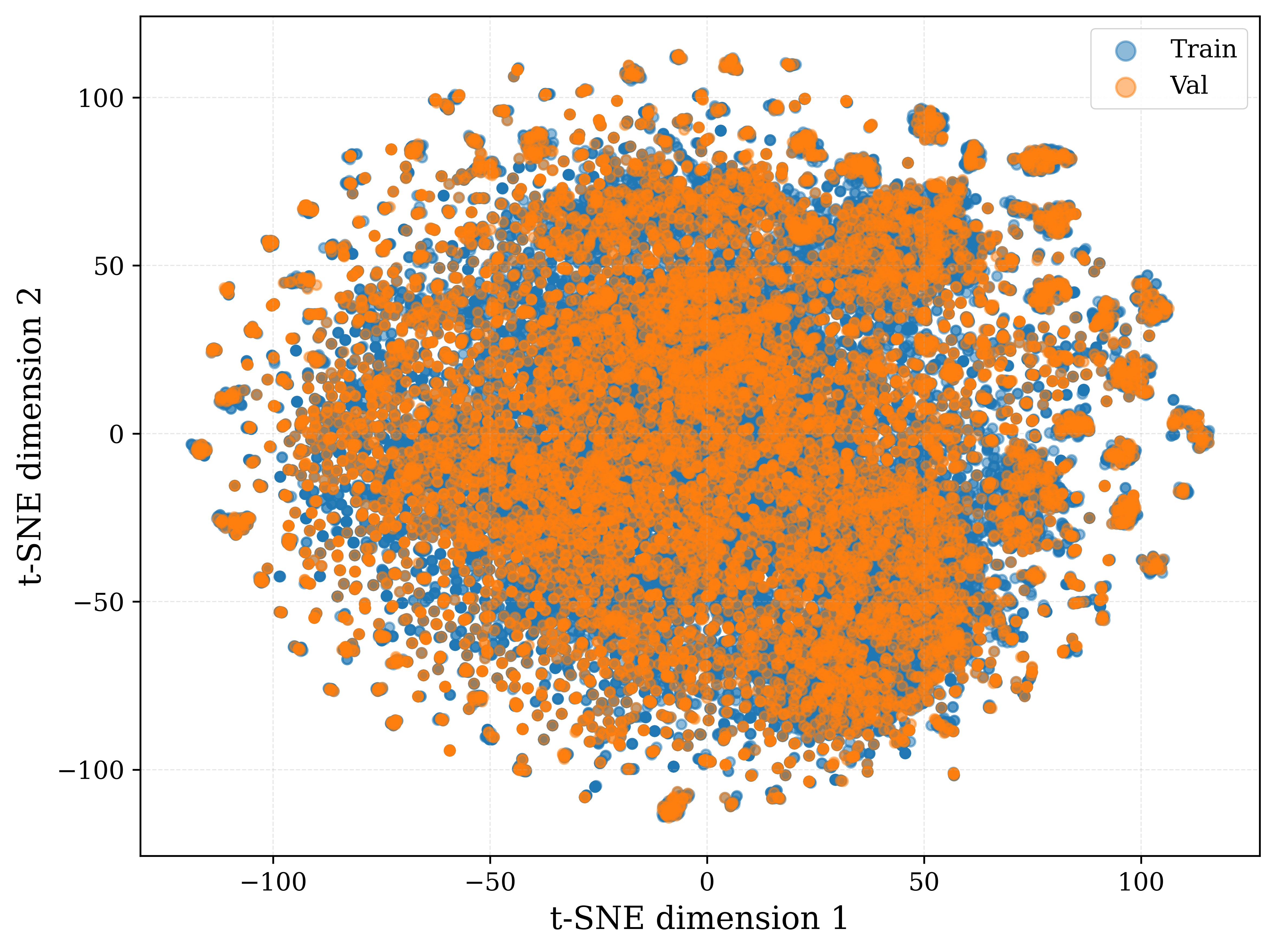}
    \caption{t-SNE visualization of DINOv3 frame embeddings for \uvhw; colored by split.}
    \label{fig:tsne-bmd-plot}
\end{figure}

To ensure spatiotemporal diversity while managing annotation costs, we applied systematic sampling:
\begin{itemize}[leftmargin=*]
    \item \textbf{Spatial coverage.} We cap the number of frames per camera to prevent location bias, ensuring representation across urban, suburban, and highway contexts. In Figure~\ref{fig:tsne-bmd-plot}, the t-SNE projection of DINOv3 embeddings forms a continuous spread with no clear clusters, unlike the cluster patterns seen for the TrafficCAM split in Figure~\ref{fig:tsne-trafficcam-plot}. This was further quantified using Vendi Scores (VS)~\cite{Friedman2022TheVS} computed on DINOv3 features: \uvhw achieves train/val VS of $225.7$/$223.0$ at $q{=}0.5$, compared to $70.6$/$43.2$ for TrafficCAM, confirming higher absolute diversity and stronger train-val consistency.
    
    \item \textbf{Temporal coverage.} The captured footage was restricted to daylight hours (06:00--18:00; local time) when cameras operated in color mode. We selected at most one frame per camera per hour, capturing traffic variations across morning, midday, and evening periods. While the collected data did not capture nighttime scenarios, the daytime coverage spanned both peak and off-peak traffic conditions, and included twilight (low-light) conditions.
\end{itemize}

This process yielded approximately 400,000 candidate images for downstream selection.

\subsubsection{Difficulty-Aware Sampling}
\label{sec:difficulty_sampling}

The annotations were performed through crowdsourced volunteers (see Section~\ref{sec:crowdsourced}). To focus the annotation effort on informative images, we adopt a difficulty-aware sampling strategy that ranks candidate frames using two image-level signals: a \textit{disagreement score} capturing inter-model uncertainty, and a \textit{composite difficulty score} capturing visual complexity. Building on evidence from active sample selection and hard-example mining for object detection~\cite{lyu2023box,wang2018ssm}, we prioritize images where detectors either strongly disagree or encounter crowded, small, or overlapping objects. These are formally defined in Appendix~\ref{app:disagree-difficulty}.

Briefly, for each image, we first compute a disagreement score, $D_i^{\mathrm{norm}}$, based on the variability in per-class box counts across multiple detectors and the maximum pairwise class-count disagreement across categories. Images where different models predict very different numbers of a class receive higher disagreement scores. This follows the idea of box-level active selection~\cite{lyu2023box}, where uncertain or conflicting boxes are more likely to reduce model error when annotated.

We also compute a composite difficulty score, $\Delta_i$, that aggregates normalized object count, average box size, pairwise IoU overlap, and the normalized disagreement signal, $\widetilde{D}_i$. Images with many objects, small average box area, high mutual overlap, and high disagreement are assigned a higher $\Delta_i$, similar in spirit to self-supervised sample mining strategies that favor hard yet informative samples~\cite{wang2018ssm}. We then sort images by $\Delta_i$ and stratify them into difficulty bands, sampling more heavily from higher bands while still retaining a fraction of easier scenes. This yields a candidate set enriched with challenging traffic conditions and contested predictions, which improves the effectiveness of both expert fine-tuning and crowdsourced annotation.

\subsection{Annotation Pipeline}
\subsubsection{Gold Dataset for Pre-annotation}
To bootstrap high-quality crowdsource annotations, we use a pre-annotation approach using a pre-trained model tuned on expert-annotated images (Figure~\ref{fig:crowdsource_workflow}).
We curate a \emph{Gold Dataset} of $3,017$ images with expert labels across all $14$ classes. This subset was split into training ($2,726$ images), validation ($906$ images), and test ($405$ images). We fine-tuned seven detector architectures initialized from COCO training: YOLOv8-{X,N}, YOLOv11-{X,N}, RT-DETRv2-X, D-FINE-X, and DAMO-YOLO-X~\cite{jocher2023yolov8, jocher2024yolov11, Lv2024_RTDETRv2, peng2025dfine, xu2022damo}.

The best-performing model, \textit{RT-DETRv2-X}, achieved an mAP@[0.5:0.95] of $70.3$ on the Gold test set, representing strong but imperfect performance, yet sufficient to accelerate annotation while requiring human verification. This model was used to generate pre-annotations for the $100,000$ images selected for crowdsourced annotations. Predictions from all seven models were retained to compute disagreement scores for quality control.

\subsubsection{Crowdsourced Annotation Workflow}\label{sec:crowdsourced}
Crowdsourced annotations were collected from volunteers through a custom web platform in a structured challenge (hackathon) format, discussed in detail in Appendix~\ref{app:crowdsourcing}.

A pilot study with 568 participants preceded the main phase to calibrate workflows, refine class definitions, and train annotators The pilot data is kept strictly separate from the final dataset.
During annotation, participants performed four tasks per image:
\begin{enumerate}[leftmargin=*, noitemsep]
    \item Verify pre-annotated bounding boxes and class labels
    \item Adjust box coordinates to tightly fit object boundaries  
    \item Add annotations for missed vehicle instances
    \item Remove false-positive detections
\end{enumerate}

Each image was independently annotated by $3$--$9$ participants (mean = 5.2 annotations/image). To enhance reliability, we dynamically prioritized images with high disagreement and difficulty scores, and inserted Gold Dataset images with known ground truth at regular intervals for in-situ quality assessment and annotator feedback.

\subsection{Consensus-Based Aggregation}\label{sec:consensus}
Multiple independent annotations per image necessitate aggregation into a single consensus ground truth. We treat bounding box localization and classification separately:

\noindent\textbf{Box localization.} Pre-annotated boxes carry persistent identifiers that annotators modify but do not delete. For each identifier, we compute the consensus box as the \textit{mean} of all submitted coordinates. New boxes added by annotators lacked identifiers; we match these across submissions using a $60\%$ IoU threshold following standard practice in crowdsourced object detection~\cite{pascalVOC2010everingham} and averaged within matched groups. This reduces individual biases while preserving localization precision.

\noindent\textbf{Class labels.} We apply \textit{majority voting} at the object level; each box receives the label selected by a plurality of annotators. Ties (occurring in $<0.5\%$ of cases) were broken uniformly, at random. This approach aligns with established crowdsourcing literature~\cite{Zhou2019, Sheshadri2013} and produces stable, interpretable labels.

To assess annotation reliability, we measured inter-annotator agreement on classification labels, obtaining $\approx87\%$ pairwise accuracy against consensus and a Fleiss' kappa of $\approx0.862$, indicating strong agreement~\cite{kappa1971fleiss, observeragreement1977landis}. Label stability converges rapidly with annotator count: $94.3\%$ consensus match at $k{=}2$ and $98.2\%$ at $k{=}5$ (the dataset mean), confirming that our $3$--$9$ annotator protocol produces stable labels. A comparison of consensus annotations against expert labels on the Gold Dataset yields a true-positive rate of $78.4\%$, false-positive rate of $5.9\%$, and false-negative rate of $15.7\%$.

\section{Comparison Protocols}
\label{sec:comparison_protocol}

We compare the detection performance of modern detectors fine-tuned on \uvhw against those fine-tuned on existing datasets with alternate perspectives or fewer samples: UA-DETRAC~\cite{Wen2020_UADETRAC} (highway views) and TrafficCAM~\cite{Deng2025_TrafficCAM} (Indian CCTV viewpoint). The primary evaluation focuses on fixed-camera CCTV images from Indian urban traffic, where we measure how well each training dataset supports accurate detection under our deployment setting. We also evaluate models fine-tuned on \uvhw on datasets with alternate viewpoints to study viewpoint generalization.

We evaluate a set of modern detectors under a unified training and evaluation protocol. All the models mentioned above are initialized from their official COCO-pretrained checkpoints and fine-tuned on the respective datasets: IDD, TrafficCAM, UA-DETRAC, and our \uvhw dataset. The detector models used are YOLOv12-S, YOLOv12-X, RT-DETRv2-X, D-FINE-X, and RF-DETR-X~\cite{Tian2025_YOLOv12, Lv2024_RTDETRv2, peng2025dfine, Robinson2025_RFDETR}. We adopt the default data processing pipelines and loss formulations provided in the official implementations of each model. For fine-tuning, we use a global batch size of $16$, resize and pad images to an input resolution of $640 \times 640$ pixels (following standard COCO-style training settings), and train for $100$ epochs. Learning-rate schedules, optimizers, weight decay, momentum, and warmup follow the configuration recommended for each model. Additional hyperparameter details are provided in Appendix~\ref{app:model-training}. All experiments are run on a mix of NVIDIA A6000 and A100 GPUs.

Our evaluation follows COCO-style object detection metrics. For each experiment, we report the mean Average Precision (mAP), averaged across IoU thresholds from $0.50$ to $0.95$ in steps of $0.05$ (mAP@$0.50$:$0.95$). We additionally provide per-class Average Precision (AP@$0.50$:$0.95$) to analyze performance on different vehicle categories. To ensure fair comparison of the \uvhw benchmark with IDD, TrafficCAM, and UA-DETRAC, we evaluate detectors only on the common vehicle classes between \uvhw and each dataset (Table~\ref{tab:common_class_mapping}); for example, our fine-grained \textit{hatchback}, \textit{sedan}, \textit{SUV}, and \textit{MUV} categories are merged into a single \textit{car} class when benchmarking against datasets that provide only a coarse \textit{car} label. IDD and TrafficCAM, which are released in segmentation format, are converted to detection format by taking tight axis-aligned bounding boxes around each instance mask. For UA-DETRAC, which is a multi-object tracking dataset, we construct a detection-only subset by sampling frames at $10$~fps and discarding temporal identities, following the methodology in~\cite{he2024enhancinguadetrac}.

\begin{table}[t]
\centering
\caption{Taxonomy-aware class mapping between \uvhw and other datasets.}
\label{tab:common_class_mapping}
\footnotesize
\setlength{\tabcolsep}{8pt} 
\renewcommand{\arraystretch}{1} 
\begin{tabular}{c l l l l}

\hline
\textbf{\#} & \textbf{\uvhw} \textit{(Ours)}& \textbf{IDD}~\cite{Varma2019_IDD} & \textbf{UA-DETRAC}\cite{Wen2020_UADETRAC} & \textbf{TrafficCAM}~\cite{Deng2025_TrafficCAM} \\\hline \hline

1 & Cycle & -- & -- & Bike \\  \hline
2 & 2-Wheeler & -- & -- & Motorbike \\  \hline
3 & 3-Wheeler & Rickshaw & -- & Auto \\  \hline
4 & Hatchback & \multirow{4}{*}{Car} & \multirow{4}{*}{Car} & \multirow{5}{*}{LMV} \\ \cmidrule(r){1-2}
5 & Sedan &  &  &  \\ \cmidrule(r){1-2}
6 & MUV &  &  &  \\ \cmidrule(r){1-2}
7 & SUV &  &  &  \\ \cmidrule(r){1-4}
8 & Van & \multirow{2}{*}{Caravan} & \multirow{2}{*}{Van} &  \\ \cmidrule(r){1-2} \cmidrule(l){5-5} 
9 & T. Traveller &  &  & -- \\ \midrule
10 & Mini-bus & \multirow{2}{*}{Bus} & \multirow{2}{*}{Bus} & \multirow{2}{*}{Bus} \\ \cmidrule(r){1-2}
11 & Bus &  &  &  \\ \midrule
12 & LCV & \multirow{2}{*}{Truck} & -- & LCV \\ \cmidrule(r){1-2} \cmidrule(l){4-5} 
13 & Truck &  & -- & Truck \\  \hline
14 & Others & -- & -- & -- \\  \hline
15 & -- & Cycle$^\dagger$ & -- & -- \\  \hline
16 & -- & 2-wheeler$^\dagger$ & -- & -- \\  \hline
17 & -- & Trailer & -- & -- \\  \hline
18 & -- & Vehicle Fallback & -- & -- \\  \hline
19 & -- & -- & Others & -- \\  \hline
20 & -- & -- & -- & Pedestrian \\  \hline
21 & -- & -- & -- & E-rickshaw \\  \hline
22 & -- & -- & -- & Tractor \\  \hline

\bottomrule
\end{tabular}

\caption*{\footnotesize
$^\dagger$ \textbf{Class Exclusion:} While \emph{Cycle} and \emph{Two-wheeler} classes appear in both \uvhw and IDD, they are excluded from cross-dataset evaluation as \uvhw's bounding boxes include both the vehicle and rider, while IDD's boxes usually cover only the vehicle, making AP comparison unreliable.
}

\end{table}

\section{Experiments and Evaluation}

We investigate cross-dataset generalization of vehicle detectors in CCTV settings. Models are trained on one of the datasets: \textbf{\uvhw}, \textbf{UA-DETRAC} or \textbf{TrafficCAM} and evaluated on the \uvhw validation split.

\subsection{Domain-Shifted Evaluation}
\textbf{UA-DETRAC vs. \uvhw.} To assess domain-shift under label intersection (car, bus, van), we compare models trained on UA-DETRAC with counterparts trained on \uvhw. In Table~\ref{tab:cross_eval_uadetrac_ours}, UA-DETRAC-trained models transfer poorly to \uvhw validation: RT-DETRv2-X trained on UA-DETRAC has \mapFiftyNinetyFive $= 0.336$ on \uvhw, vs. $0.838$ when trained on \uvhw; analogous gaps appear for RF-DETR (UA-DETRAC: $0.232$ vs. \uvhw: $0.795$) and D-FINE (\uvhw: $0.828$). Although UA-DETRAC contains $1.21M$ bounding boxes, it represents only $8.2K$ unique vehicles due to the dataset's focus on tracking. This produces substantial temporal redundancy and limited appearance diversity. Thus, models trained on UA-DETRAC struggle with the heterogeneous, fine-grained categories and broader camera distribution present in \uvhw.

\subsection{In-Domain Evaluation}
\textbf{TrafficCAM vs. \uvhw.} We repeat the analysis with TrafficCAM, evaluating over the intersecting classes of {Auto, LMV, LCV, Motorbike and Truck}. Despite having a CCTV viewpoint, its labeled subset is small ($4.3K$ frames). As shown in Table~\ref{tab:cross_eval_trafficcam_ours}, models trained on TrafficCAM transfer better relative to UA-DETRAC but remain clearly below \uvhw-trained models: RT-DETRv2-X trained on TrafficCAM reaches \mapFiftyNinetyFive~$= 0.474$ on \uvhw, compared to $0.798$ when trained on \uvhw; RF-DETR and D-FINE show similar gaps. The shortfall is consistent with scale and annotation consistency: \uvhw's broader camera coverage and consensus-driven labels yield better separation even when evaluation is restricted to the common label set. We believe that our disagreement-driven image selection also provides a more diverse, difficult, and informative representation of vehicles in the wild.

Across architectures and intersecting taxonomies, detectors trained on structured or smaller CCTV corpora exhibit limited transfer to new environments, underscoring the value of large-scale, CCTV-native datasets like \uvhw for ITS deployment.

\subsection{Mixed-Dataset Evaluation}
To further isolate the contribution of dataset composition, we conduct ablation studies by augmenting UA-DETRAC and TrafficCAM with subsets drawn from \uvhw. We construct two mixed testing regimes: \textit{Random Sampling (MRS)}, where additional frames are sampled uniformly from the dataset pairs, and \textit{High-Difficulty Selection (MHD)}, where frames are chosen based on disagreement metrics or detection complexity to emphasize challenging examples. Both mixtures preserve the label intersection used in prior cross-evaluation experiments.
Tables~\ref{tab:cross_eval_trafficcam_ours} and~\ref{tab:cross_eval_uadetrac_ours} report the mixed-dataset results under the MHD and MRS settings. Across both evaluations, models trained on \uvhw show higher values of \mapFiftyNinetyFive{} and \mapFifty{} when evaluated on these splits. The performance trend suggests that the wider class coverage and diverse conditions in \uvhw support better generalization.

\begin{table}[t]
\centering
\caption{Cross evaluation - UA-DETRAC and \uvhw}
\label{tab:cross_eval_uadetrac_ours}
\footnotesize

\begin{tabular}{ccc ccc}

\toprule
\textbf{Trained On}& \textbf{Evaluated On$^\dagger$} & \textbf{Model}       
& \textbf{\mapFiftyNinetyFive} & \textbf{\mapSeventyFive} & \textbf{\mapFifty} \\
\hline\hline

\multirow{5}{*}{UA-DETRAC} & \multirow{5}{*}{MHD} 
& RT-DETRv2 X & 0.56 & 0.65 & 0.71 \\
& & D-FINE       & 0.49 & 0.58 & 0.64 \\
& & RF-DETR X    & 0.51 & 0.59 & 0.66 \\
& & YOLOv12 S    & 0.38 & 0.44 & 0.53 \\
& & YOLOv12 X    & 0.34 & 0.39 & 0.46 \\

\hline

\multirow{5}{*}{\uvhw} & \multirow{5}{*}{MHD} 
& RT-DETRv2 X & 0.64 & 0.74 & 0.83 \\
& & D-FINE       & \textbf{0.65} & \textbf{0.76} & \textbf{0.84} \\
& & RF-DETR X    & 0.64 & 0.75 & \textbf{0.84} \\
& & YOLOv12 S    & 0.42 & 0.47 & 0.59 \\
& & YOLOv12 X    & 0.29 & 0.29 & 0.51 \\

\hline\hline

\multirow{5}{*}{UA-DETRAC} & \multirow{5}{*}{MRS}
& RT-DETRv2 X & 0.57 & 0.66 & 0.73 \\
& & D-FINE       & 0.51 & 0.60 & 0.66 \\
& & RF-DETR X    & 0.52 & 0.61 & 0.67 \\
& & YOLOv12 S    & 0.39 & 0.46 & 0.55 \\
& & YOLOv12 X    & 0.35 & 0.40 & 0.48 \\

\hline

\multirow{5}{*}{\uvhw} & \multirow{5}{*}{MRS}
& RT-DETRv2 X & 0.63 & 0.74 & 0.82 \\
& & D-FINE       & \textbf{0.64} & \textbf{0.75} & \textbf{0.84} \\
& & RF-DETR X    & \textbf{0.64} & \textbf{0.75} & \textbf{0.84} \\
& & YOLOv12 S    & 0.41 & 0.47 & 0.59 \\
& & YOLOv12 X    & 0.28 & 0.29 & 0.50 \\

\bottomrule
\end{tabular}
\caption*{\footnotesize
$^\dagger$ \textbf{MHD}: Mixed; High-Difficulty. \textbf{MRS}: Mixed; Randomly Sampled.
}
\end{table}

\begin{table}[t]
\centering

\caption{Cross evaluation - TrafficCAM and \uvhw}
\label{tab:cross_eval_trafficcam_ours}
\footnotesize
\begin{tabular}{ccc ccc}
\toprule
\multicolumn{1}{l}{\textbf{Trained On}} & \multicolumn{1}{l}{\textbf{Evaluated On$^\dagger$}} & \textbf{Model} & \textbf{\mapFiftyNinetyFive} & \textbf{\mapSeventyFive} & \textbf{\mapFifty} \\
\hline\hline

\multirow{5}{*}{TrafficCAM} & \multirow{5}{*}{MHD} 
& RT-DETRv2 X & 0.48 & 0.53 & 0.65 \\
& & D-FINE       & 0.48 & 0.53 & 0.65 \\
& & RF-DETR X    & 0.40 & 0.45 & 0.58 \\
& & YOLOv12 S    & 0.23 & 0.25 & 0.35 \\
& & YOLOv12 X    & 0.21 & 0.22 & 0.29 \\

\hline

\multirow{5}{*}{\uvhw} & \multirow{5}{*}{MHD} 
& RT-DETRv2 X & \textbf{0.56} & \textbf{0.61} & \textbf{0.70} \\
& & D-FINE       & 0.55 & 0.60 & 0.69 \\
& & RF-DETR X    & 0.52 & 0.57 & 0.68 \\
& & YOLOv12 S    & 0.35 & 0.38 & 0.51 \\
& & YOLOv12 X    & 0.25 & 0.24 & 0.44 \\

\hline\hline

\multirow{5}{*}{TrafficCAM} & \multirow{5}{*}{MRS} 
& RT-DETRv2 X & 0.49 & 0.54 & \textbf{0.67} \\
& & D-FINE       & \textbf{0.50} & \textbf{0.55} & \textbf{0.67} \\
& & RF-DETR X    & 0.42 & 0.46 & 0.59 \\
& & YOLOv12 S    & 0.26 & 0.28 & 0.37 \\
& & YOLOv12 X    & 0.23 & 0.25 & 0.32 \\

\hline

\multirow{5}{*}{\uvhw} & \multirow{5}{*}{MRS} 
& RT-DETRv2 X & 0.49 & 0.54 & 0.64 \\
& & D-FINE       & 0.49 & 0.54 & 0.64 \\
& & RF-DETR X    & 0.46 & 0.51 & 0.63 \\
& & YOLOv12 S    & 0.31 & 0.33 & 0.45 \\
& & YOLOv12 X    & 0.21 & 0.21 & 0.38 \\

\bottomrule
\end{tabular}

\caption*{\footnotesize
$^\dagger$ \textbf{MHD}: Mixed; High-Difficulty. \textbf{MRS}: Mixed; Randomly Sampled.
}

\end{table}

\begin{figure}[t]
    \centering
    \includegraphics[width=1\columnwidth]{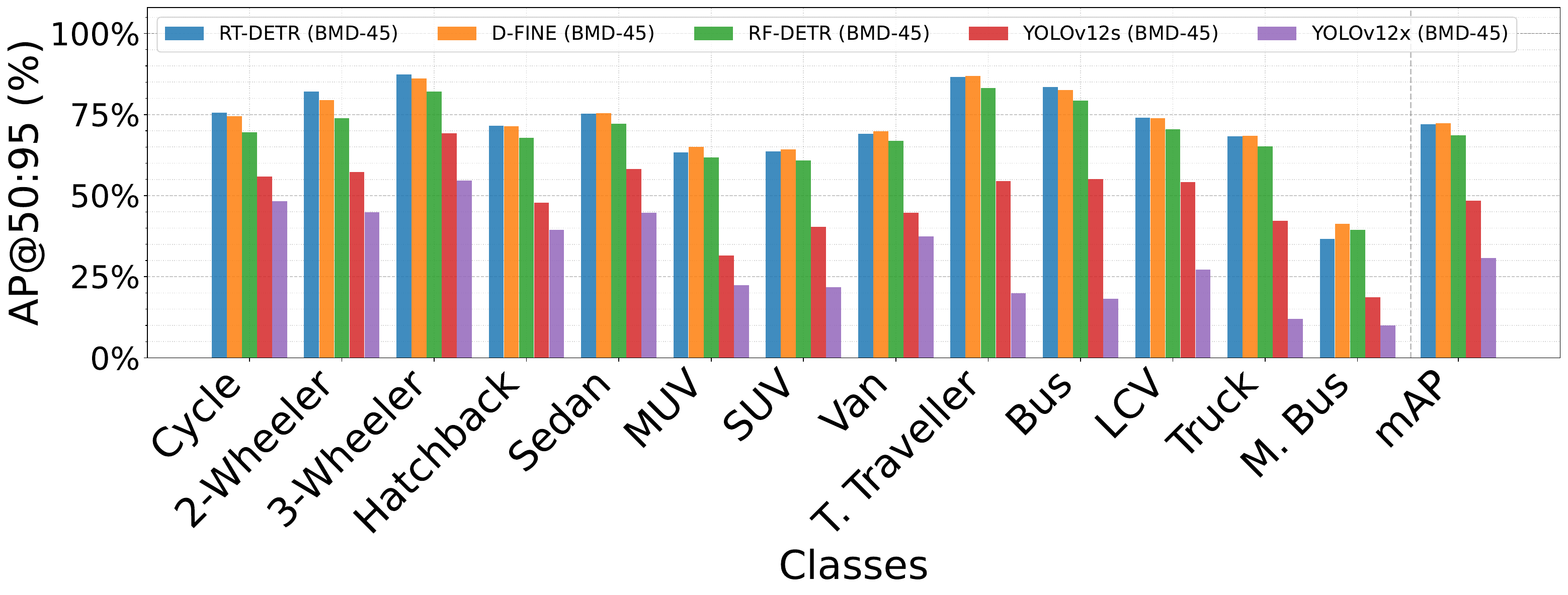}
    \caption{AP@50:95 distribution for selected models trained on \uvhw dataset and inferred on \uvhw val set.}
    \label{fig:model_evaluation_hackathon}
\end{figure}

\section{Conclusion}

This work introduces \uvhw, a large-scale fixed-camera vehicle detection dataset from Bengaluru that targets heterogeneous and unstructured urban traffic, and examines how current datasets and detectors perform in this setting. By combining difficulty-aware sampling, a multi-stage crowdsourced annotation pipeline, and consensus-based aggregation, \uvhw provides $45,986$ images with $481,947$ bounding boxes across $14$ fine-grained vehicle classes commonly found in Indian cities. Our comparison protocols and cross-dataset evaluations show that models trained on existing datasets (IDD, UA-DETRAC, TrafficCAM), and open-set methods like GroundingDINO, achieve only moderate mAP values on our expert-annotated reference set. These same architectures trained on \uvhw obtain higher accuracy on held-out CCTV data and transfer better to other fixed-camera benchmarks. These findings suggest that scale alone is insufficient; viewpoint variety, regional vehicle categories, and scene difficulty are equally important for CCTV-based ITS deployment. A key limitation of the present dataset is its focus on a single city and daytime conditions, which we hope to address in the future.

\section*{Ethics Statement}
\label{sec:ethics}

All data used in this work were collected from publicly deployed traffic cameras operated by the Bengaluru Traffic Police (BTP) under appropriate institutional oversight. The dataset release and associated research were approved by our Institutional Review Board (IRB). Before public release, we applied standard anonymization procedures: vehicle license plates, human faces and camera text overlays were blurred, consistent with best practices~\cite{frome2009streetview, cityscapes_scripts, Waymo_FAQ, Mapillary_Privacy, Varma2019_IDD}; no personally identifiable information or sensitive metadata is present. Access to original, unblurred images was restricted to authorized annotators. 

A small additional set of images, collected under non-disclosure terms for preliminary experiments and protocol design, is not publicly shareable; all quantitative results reported in this paper and all SOTA comparisons are based solely on the larger publicly available test set that is being released. 

\section{Acknowledgments}
We thank the Bengaluru Traffic Police (BTP) and the Bengaluru Police for providing access to the Safe City camera data from which the image datasets used for this release were derived. We thank Capital One for sponsoring the prizes for the Urban Vision Hackathon 2025 competition. We thank IISc's AI and Robotics Technology Park (ARTPARK) and Centre for infrastructure, Sustainable Transportation and Urban Planning (CiSTUP) for funding the annotation and model training efforts, and Kotak IISc AI-ML Centre (KIAC) for providing the GPU resources required to train the models. We acknowledge the outreach support provided by the ACM India Council and IEEE India Council to help encourage chapter volunteers to participate in Urban Vision Hackathon 2025.

%% ARXIV STYLE
\bibliographystyle{plain}
\bibliography{main.bib}

\input{appendix.tex}

\end{document}

%% file: appendix.tex
\clearpage
\appendix

\section{Release Notes}
\label{app:release-notes}
The datasets and models are posted on Huggingface under \url{https://huggingface.co/iisc-aim/}. The datasets are under \url{https://huggingface.co/datasets/iisc-aim/BMD-45} while models are under \url{https://huggingface.co/iisc-aim/BMD-45}.

\subsection{Datasets}
The folder structure on Huggingface for the datasets present under \url{https://huggingface.co/datasets/iisc-aim/BMD-45/tree/main} is as follows:
\begin{itemize}
    \item \texttt{BMD-45-Train/}: This folder has the $\sim$78\% of \uvhw data used for training.
    \begin{itemize}
        \item \texttt{images\_000/} through \texttt{images\_007/}: Training images organized into 8 sub-folders for convenience.
        \item \texttt{images\_000/*}: Actual training images with filenames such as \texttt{41.png, 47.png, ...} that are unique across the entire dataset.
        \item \texttt{images\_001/*}, \ldots, \texttt{images\_007/*}: Additional sub-folders following the same structure.
        \item \texttt{\_annotations.coco.json}: Majority Voting consensus annotations for the training images, provided in COCO JSON format.
        \item \texttt{metadata.jsonl}: HuggingFace ImageFolder annotations (one JSON line per image).
    \end{itemize}
    \item \texttt{BMD-45-Val/}: This folder has the $\sim$22\% of \uvhw data used for validation.
    \begin{itemize}
        \item \texttt{images\_000/} through \texttt{images\_002/}: Validation images organized into 3 sub-folders for convenience.
        \item \texttt{images\_000/*}: Validation images. All filenames are globally unique across both training and validation sets.
        \item \texttt{images\_001/*}, \texttt{images\_002/*}: Additional sub-folders following the same structure.
        \item \texttt{\_annotations.coco.json}: Majority Voting consensus annotations for the validation images, provided in COCO JSON format.
        \item \texttt{metadata.jsonl}: HuggingFace ImageFolder annotations (one JSON line per image).
    \end{itemize}
    \item \texttt{LICENSE}: License file mentioning the \textit{Creative Commons Attribution 4.0 International License}~\footnote{\url{https://creativecommons.org/licenses/by/4.0/}} under which the data is being released.
\end{itemize}

\subsection{Models}
The folder structure on Huggingface for the models present under \url{https://huggingface.co/iisc-aim/BMD-45/tree/main} is as follows:

\begin{itemize}
    \item \texttt{bmd\_classes.txt}: Text file with the list of 13 BMD-45 object classes used for training, with one class per line.
    \item \texttt{configs/}: Configuration files defining hyperparameters and architecture details used during model training, organized by model family.
    \begin{itemize}
        \item \texttt{YOLOv12-S/}: Configuration for the S-sized YOLOv12 model.
        \begin{itemize}
            \item \texttt{config.yaml}: Training hyperparameters.
            \item \texttt{data.yaml}: Dataset paths and class names.
        \end{itemize}
        \item \texttt{YOLOv12-X/}: Configuration for the X-sized YOLOv12 model.
        \begin{itemize}
            \item \texttt{config.yaml}: Training hyperparameters.
            \item \texttt{data.yaml}: Dataset paths and class names.
        \end{itemize}
        \item \texttt{RT-DETRv2/}: Configuration for the RT-DETRv2 model.
        \begin{itemize}
            \item \texttt{bmd-45-dataset.yaml}: Dataset configuration.
            \item \texttt{rtdetrv2\_r101vd\_6x\_bmd-45.yaml}: Model and training configuration.
        \end{itemize}
        \item \texttt{RF-DETR/}: Configuration for the RF-DETR model.
        \begin{itemize}
            \item \texttt{config.yaml}: Training hyperparameters.
        \end{itemize}
        \item \texttt{D-FINE/}: Configuration for the D-FINE model.
        \begin{itemize}
            \item \texttt{bmd-45-dataset.yaml}: Dataset configuration.
            \item \texttt{dfine\_hgnetv2\_x\_bmd-45.yaml}: Model and training configuration.
        \end{itemize}
    \end{itemize}
    \item \texttt{weights/}: Directory containing trained model weights organized by model family.
    \begin{itemize}
        \item \texttt{YOLOv12-S/}: Weights for the S-sized YOLOv12 model variant.
        \begin{itemize}
            \item \texttt{best.pt}: Trained weights in PyTorch format.
        \end{itemize}
        \item \texttt{YOLOv12-X/}: Weights for the X-sized YOLOv12 model variant.
        \begin{itemize}
            \item \texttt{best.pt}: Trained weights in PyTorch format.
        \end{itemize}
        \item \texttt{RT-DETRv2/}: Weights for the RT-DETRv2 model variant.
        \begin{itemize}
            \item \texttt{best.pth}: Trained weights in PyTorch format.
        \end{itemize}
        \item \texttt{RF-DETR/}: Weights for the RF-DETR model variant.
        \begin{itemize}
            \item \texttt{checkpoint\_best\_total.pth}: Trained weights in PyTorch format.
        \end{itemize}
        \item \texttt{D-FINE/}: Weights for the D-FINE model variant.
        \begin{itemize}
            \item \texttt{best\_stg1.pth}: Trained weights in PyTorch format.
        \end{itemize}
    \end{itemize}
    \item \texttt{usage/}: Folder for example scripts demonstrating model loading and inference.
\end{itemize}

\section{Dataset Details}
\label{app:dataset-details}
We present additional details about the \uvhw dataset and its annotation process in this section.

\subsection{\uvhw Vehicle Classes}
\label{app:vehicle-classes}
\begin{figure}[hbt!]
    \centering
    \includegraphics[width=1\columnwidth]{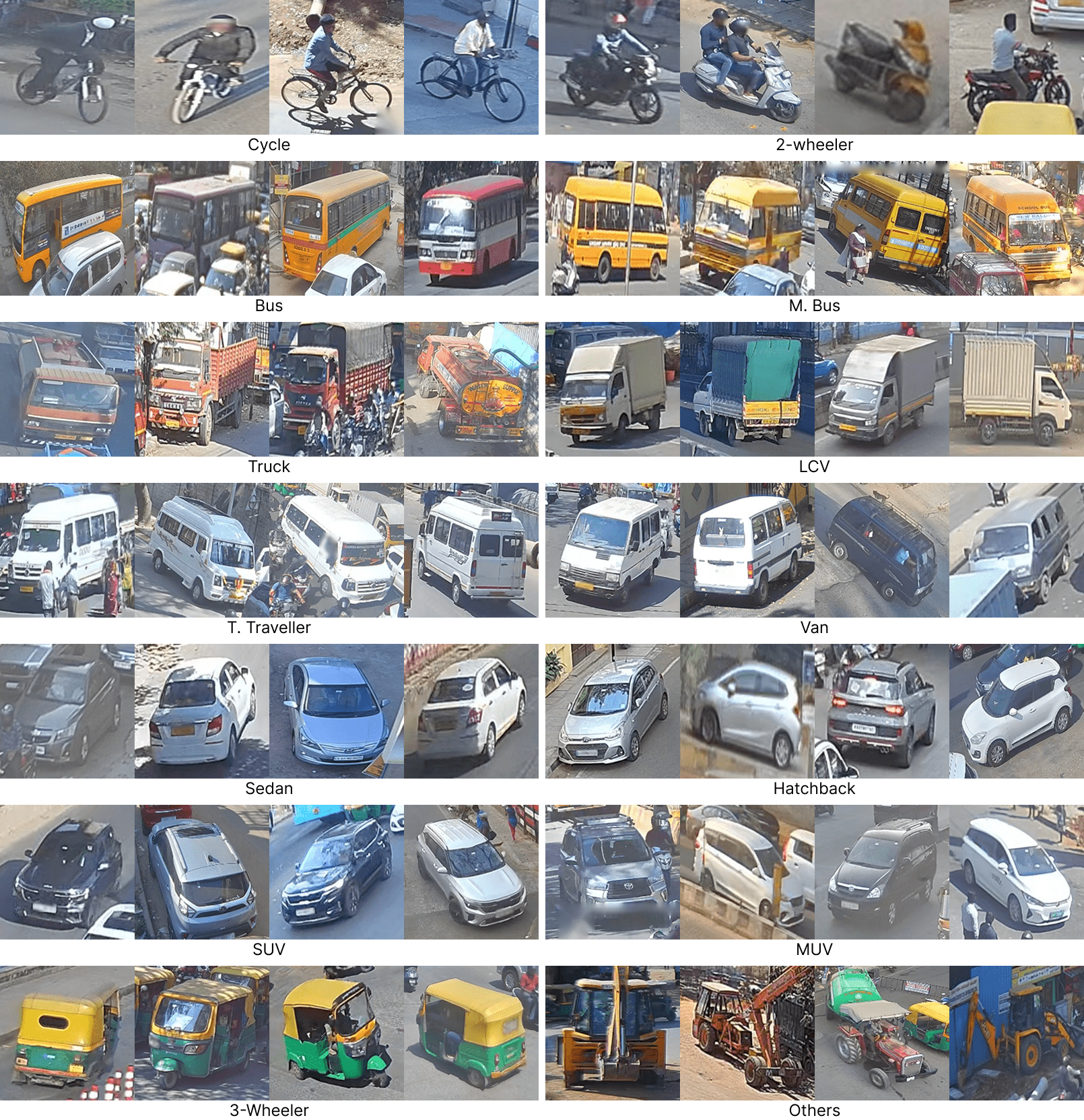}
    \caption{Example cropped images of each of 14 classes in the \uvhw dataset.}
    \label{fig:example_classes}
\end{figure}

As mentioned in the main text, we focus on $14$ fine-grained vehicle classes that reflect the diversity of India's vehicle fleet. Detailed descriptions of each vehicle class are provided in Table~\ref{tab:vehicle_category_description} of the Appendix, and cropped samples of the vehicles from each class are shown in Figure~\ref{fig:example_classes}.

\begin{table}[hbt!]
\centering
\small
\caption{Vehicle Classes and Descriptions}
\label{tab:vehicle_category_description}
\begin{tabular}{lL{10cm}}
\toprule
\textbf{Class} & \textbf{Description} \\
\midrule
\midrule
\textbf{Cycle} & Non-motorized, manually pedaled vehicles including geared, non-geared, women's, and children's cycles. Bounding boxes include both the vehicle and rider.\\
\midrule
\textbf{2-Wheeler} & Motorbikes and scooters for single or double riders. Bounding boxes include both vehicle and rider.\\
\midrule
\textbf{3-Wheeler} & (i.e., Auto-rickshaw) Compact vehicles with one front wheel and two rear wheels, featuring a covered passenger cabin.\\
\midrule
\textbf{Hatchback} & Small passenger cars without a protruding rear boot/trunk.\\
\midrule
\textbf{Sedan} & Passenger cars with a low-slung design and a separate protruding rear trunk/boot.\\
\midrule
\textbf{MUV} & (i.e., Multi-Utility Vehicle) Large vehicles with three seating rows, combining passenger and cargo functionality.\\
\midrule
\textbf{SUV} & (i.e., Sport Utility Vehicle) Car-like vehicles with high ground clearance, a sturdy body, and no protruding boot.\\
\midrule
\textbf{Van} & Medium-sized vehicles for transporting goods or people, typically with a flat front and sliding side doors. Smaller than Tempo Travellers.\\
\midrule
\textbf{T. Traveller} & (i.e., Tempo Traveller) Medium-sized passenger vans with tall roofs and side windows. Larger than vans but smaller than minibuses, with a protruding front.\\
\midrule
\textbf{M. Bus} & (i.e., Mini Bus) Shorter, compact buses with fewer seats. Larger than a Tempo Traveller, often featuring a flat front.\\
\midrule
\textbf{Bus} & Large passenger vehicles used for public or private transport, including office shuttles and intercity buses.\\
\midrule
\textbf{LCV} & (i.e., Light Commercial Vehicle) Lightweight goods carriers used for short to medium distance transport.\\
\midrule
\textbf{Truck} & Heavy goods carriers with a front cabin and a rear cargo compartment.\\
\midrule
\textbf{Other} & Vehicles not covered in other classes, including agricultural, specialized, or unconventional designs.\\
\bottomrule
\end{tabular}
\end{table}

\subsection{Additional Dataset Statistics}
\label{app:distributions}

\begin{figure}[hbt!]
    \centering
    \includegraphics[width=0.8\columnwidth]{
        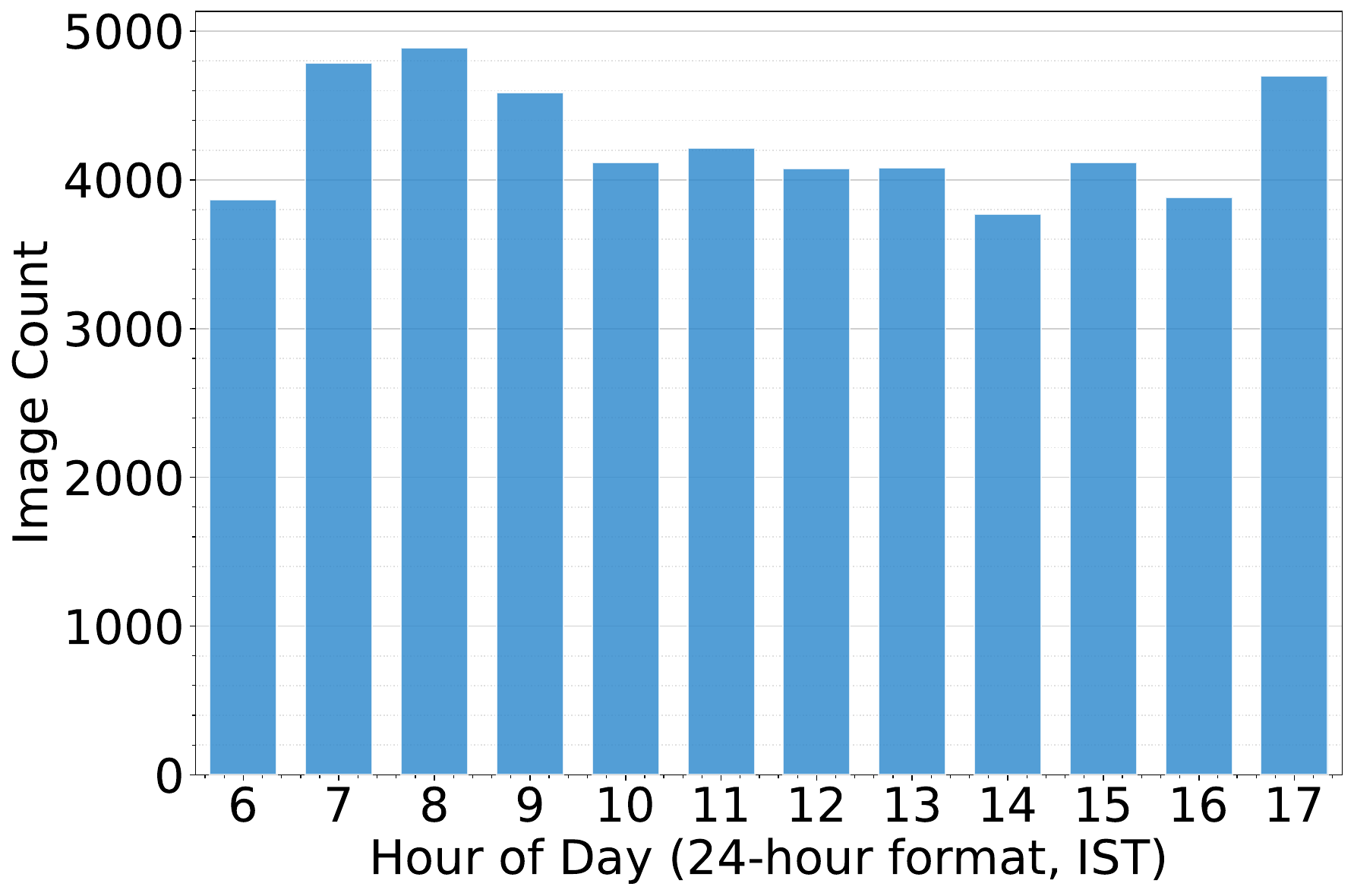
    }
    \caption{Time of day distribution of images in \uvhw across 25 days.}
    \label{fig:hour_distribution}
\end{figure}

\begin{figure}[hbt!]
    \centering
    \includegraphics[width=0.8\columnwidth]{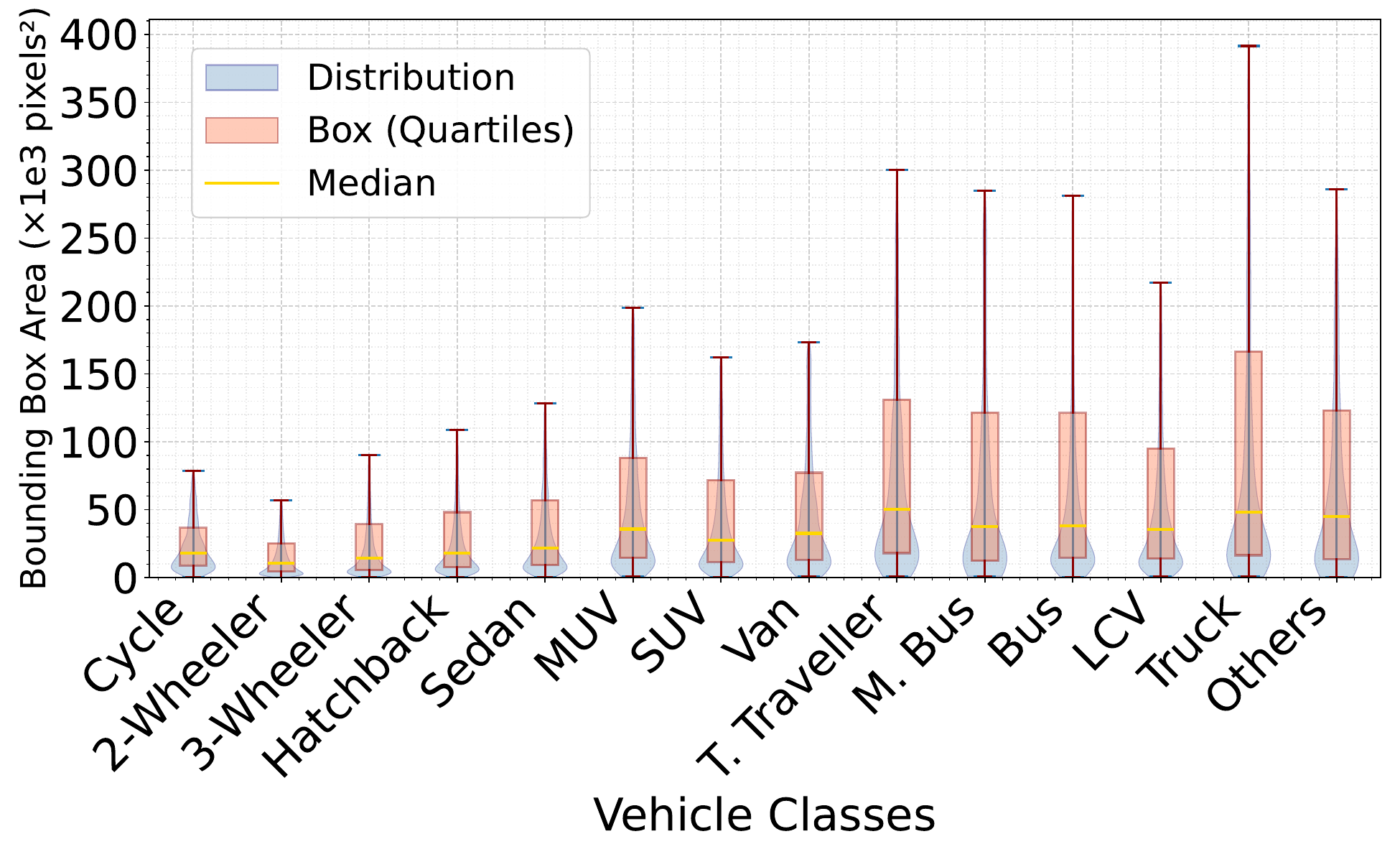}
    \caption{Distribution of bounding box area across all the classes in \uvhw}
    \label{fig:area_class_dist}
\end{figure}

We provide additional statistics to further characterize the \uvhw dataset. Figure~\ref{fig:hour_distribution} illustrates the distribution of image timestamps across daylight hours, complementing the temporal properties discussed in the main paper. Figure~\ref{fig:area_class_dist} shows the distribution of bounding-box areas across all vehicle classes, summarizing the range of object scales present in our fixed-camera CCTV views. These plots offer a more comprehensive view of the dataset’s diversity and scene characteristics.

\section{Disagreement and Image Difficulty}
\label{app:disagree-difficulty}

\paragraph{Notation.} Let $M$ be the number of detectors/models (or annotators) used for comparison and $C$ the number of classes. For a given image, let $c_{m,i}$ denote the count of bounding boxes of class $i$ predicted by model $m$ ($m\in\{1,\dots,M\}$, $i\in\{1,\dots,C\}$). Define $B_m=\sum_{i=1}^C c_{m,i}$ as the total bounding-box count produced by model $m$. We use $i$ as an image index where required; when ambiguity is possible we write $c_{m,i}^{(img)}$ or $D_i$ for the image-level disagreement score of image $i$.

\subsection{Disagreement Score}
The disagreement score captures four complementary aspects of inter-model variability.
\begin{itemize}
 
  \item \textbf{Per-class Count Disagreement.} For each class $i$ compute the standard deviation of counts across models:
  \begin{equation}
  \begin{split}
    \sigma(c_i)=\sqrt{\frac{1}{M}\sum_{m=1}^M \big(c_{m,i}-\overline{c_i}\big)^2} \\
    % \qquad
    \overline{c_i}=\frac{1}{M}\sum_{m=1}^M c_{m,i}
  \end{split}
  \end{equation}
  Summing across classes yields the per-image class-count disagreement:
  \begin{equation}
    N_{dci}=\sum_{i=1}^C \sigma(c_i)
  \end{equation}
  This term measures how much the models disagree on counts for each class (e.g., some models see three three-wheelers while others see one).

  \item \textbf{Maximum Pairwise Class-count Disagreements.} For each class $i$ count how many model pairs disagree in their class counts:
  \begin{equation}
    D_i = \sum_{m=1}^{M-1}\sum_{n=m+1}^{M} \mathbb{I}\big(c_{m,i} \neq c_{n,i}\big)
  \end{equation}
  Then take the worst (maximum) across classes:
  \begin{equation}
    M_{mdi} = \max_{i\in\{1,\dots,C\}} D_i
  \end{equation}
  $M_{mdi}$ highlights the single class with the largest pairwise disagreement and emphasizes hard, contested categories.
\end{itemize}

We combine the main components into a compact per-image disagreement score:
\begin{equation}
  D_i = N_{dci} + M_{mdi}
\end{equation}
which balances aggregate count variance with the worst-case per-class pairwise disagreement. To make scores comparable across the dataset, we normalize:
\begin{equation}
  D_i^{\mathrm{norm}} = \frac{D_i - D_{\min}}{D_{\max} - D_{\min}} \times 100
\end{equation}
where $D_{\min}$ and $D_{\max}$ are the observed minimum and maximum $D_i$ values. All component quantities used in selection (e.g., $V_{uci},\,V_{nbi},\,N_{dci},\,M_{mdi}$) are retained for analysis and can be inspected individually when diagnosing why a particular image is contentious.

\subsection{Difficulty Score}
While disagreement measures \emph{inter-model uncertainty} (useful to prioritize images where detectors disagree), it does not by itself quantify visual complexity. To ensure annotator workload was balanced and to construct zone-wise difficulty progression in the gamified challenge, we computed a complementary image-level \emph{difficulty score} that captures intrinsic visual factors (object count, scale, density, overlap) together with model disagreement.

\paragraph{Definitions.} For a given image of resolution $H\times W$ with $N_{\mathrm{bboxes}}$ detected or annotated boxes $B_1,\dots,B_{N_{\mathrm{bboxes}}}$ (each box $B_j$ has width $w_j$ and height $h_j$), we compute the following normalized components:
\begin{itemize}
  \item \textbf{Bounding-box Count:} 
  \begin{equation}
    M_{\text{bbox\_count}} = N_{\mathrm{bboxes}}
  \end{equation}
  normalized by a dataset maximum $M_{\text{bb\_max}}$:
  \begin{equation}
    \tilde{C} = \frac{M_{\text{bbox\_count}}}{M_{\text{bb\_max}}}\in[0,1]
  \end{equation}

  \item \textbf{Average Box Size:} the mean relative area
  \begin{equation}
    M_{\text{bbox\_size}} = \frac{1}{HW}\cdot\frac{1}{N_{\mathrm{bboxes}}}\sum_{j=1}^{N_{\mathrm{bboxes}}} w_j h_j
  \end{equation}
  and we use its complement 
  \begin{equation}
    (1 - M_{\text{bbox\_size}})\in[0,1]
  \end{equation}
  so that smaller average objects increase difficulty.

  \item \textbf{Bounding-box Density:} total box area fraction
  \begin{equation}
    M_{\text{bbox\_density}} = \frac{1}{HW}\sum_{j=1}^{N_{\mathrm{bboxes}}} w_j h_j
  \end{equation}
  clipped or normalized into $[0,1]$ (we use $\min(1, M_{\text{bbox\_density}})$)

  \item \textbf{Class Diversity:} 
  \begin{equation}
    M_{\text{class\_count}} = |\{\text{unique classes in image}\}|
  \end{equation}
  normalized by the maximum number of classes $M_{\text{max\_classes}}$:
  \begin{equation}
    \tilde{K} = \frac{M_{\text{class\_count}}}{M_{\text{max\_classes}}}\in[0,1]
  \end{equation}

  \item \textbf{Average IoU Overlap:} for each unordered box pair $(B_p,B_q)$ define
  \begin{equation}
    \mathrm{IoU}(B_p,B_q)=\frac{|B_p\cap B_q|}{|B_p\cup B_q|}
  \end{equation}
  and the mean pairwise overlap

  \begin{equation}
  \begin{split}
    M_{\mathrm{iou\_overlap}} = \frac{1}{N_{\mathrm{pairs}}}\sum_{p<q}\mathrm{IoU}(B_p,B_q) \\
  \quad N_{\mathrm{pairs}}=\binom{N_{\mathrm{bboxes}}}{2}
  \end{split}
  \end{equation}
  
  High $M_{\mathrm{iou\_overlap}}$ indicates occlusion and crowding.
  
  \item \textbf{Model Disagreement (Normalized):} we reuse the disagreement score above and scale it to $[0,1]$:
  \begin{equation}
    \widetilde{D}_i = \frac{D_i^{\mathrm{norm}}}{100}\in[0,1]
  \end{equation}
\end{itemize}

\subsubsection{Composite Difficulty Score} The image difficulty $\Delta_i$ is a weighted sum of normalized components:
\begin{equation}
  \Delta_i = \tilde{C}\;+\; (1 - M_{\text{bbox\_size}})\;+\; \widetilde{D}_i\;+\; M_{\mathrm{iou\_overlap}}
\end{equation}
here all components are pre-normalized to $[0,1]$. Optionally, one can include $\tilde{K}$ (class diversity) or $M_{\text{bbox\_density}}$ as additional terms if finer control is required. Finally, as with disagreement, $\Delta_i$ can be rescaled to $[0,100]$ for presentation.

\section{Crowdsourcing Annotations}
\label{app:crowdsourcing}

Large-scale expert annotation of CCTV imagery is prohibitively expensive; therefore, we adopted a controlled crowdsourcing strategy to obtain high-quality labels at scale. A total of 568 volunteer participants contributed annotations through a custom web platform during a 5-week online challenge.

As described in the main text, to reduce workload and improve consistency, each image was presented with \emph{pre-annotations} produced by the RT-DETRv2-X model fine-tuned on our Gold Dataset. Participants verified and corrected these predictions by adjusting bounding boxes, editing class labels, and adding or removing instances. Images were selected for annotation based on model disagreement scores to focus human effort on diverse and informative cases, while difficulty was varied to mitigate fatigue.

Quality was monitored using \emph{Gold} images inserted uniformly throughout the workflow. Participants were presented a mixture of known ground-truth (gold) and unknown (non-gold) images in randomized order and they were not informed of the presence of the gold images. These gold images provided per-participant accuracy estimates and enabled automatic detection of low-quality submissions.

Each image received between $3$ and $9$ independent annotations (mean $\approx 5$), allowing both coverage and redundancy. Annotation assignment ensured that participants did not see repeated images and that the overall distribution balanced load with dataset breadth. As detailed in \S\ref{sec:consensus}, the resulting multi-annotator submissions were subsequently aggregated into a single consensus using IoU-based box matching for localization and majority voting for class labels.

\section{Comparison with Other Datasets}
\label{app:comparison}

\subsection{Vehicle Perception Datasets}
\label{app:compare-other-datasets}

\begin{table}[t]
\centering
\small
\caption{Comparison of major vehicle detection and tracking datasets.}
\label{tab:app-vehicle-datasets-comp}
\resizebox{\linewidth}{!}{
\begin{tabular}{lllcccccc}
\toprule
\textbf{Dataset} & \textbf{Venue (Year)} & \textbf{Task$^\dagger$} & \textbf{PoV$^\ddagger$} & \textbf{\#Frames} & \textbf{\#Annotations} & \textbf{\#Veh.~classes} & \textbf{\#Cameras$^\diamond$} & \textbf{Location} \\
\midrule
\midrule
\multicolumn{9}{c}{\textbf{Moving-camera}} \\
\midrule
KITTI~\cite{Geiger2012_KITTI} & CVPR 2012 & D, M & E & 15k & 80k & 3 & - & DE \\
Cityscapes~\cite{Cordts2015_Cityscapes} & CVPR 2016 & S & E & 25k & 65.4k & 8 & - & DE \\
UAV-DT~\cite{du2018uavdt} & ECCV 2018 & D, M & T & 80k & $\approx841.5k$ & 3 & - & CN \\
VisDrone~\cite{visdrone} & ECCV 2018 & D, M & T & 262k & 2.6M & 8 & - & CN \\
IDD~\cite{Varma2019_IDD} & WACV 2019 & D, S & E & 10k & $\ 111.3k$ & 9 & - & IN \\
BDD100K~\cite{Yu2020_BDD100K} & CVPR 2020 & D, S, M & E & 10k & 3.3M & 5 & - & US \\
\midrule
\multicolumn{9}{c}{\textbf{Fixed-camera}} \\
\midrule
CityCam~\cite{Zhang2017_CityCam} & CVPR 2017 & D & FC & 60k & 900k & 10 & 212 & US \\
CityFlow~\cite{Tang2019_CityFlow} & CVPR 2019 & D, M & F & 117k & 230k & 9 & 40 & US \\
UA-DETRAC~\cite{Wen2020_UADETRAC} & CVIU 2020 & D, M & F & 140k & $1.21M$ & 4 & 24$^\ast$ & CN \\
TrafficCAM~\cite{Deng2025_TrafficCAM} & T-ITS 2025 & S & FC & 4.3k & $\approx 84.2k$ & 9 & NA & IN \\
\textbf{\uvhw (Our)} & -- & D & FC & 45k & $481.9k$ & 14 & 3679 & IN \\
\bottomrule
\end{tabular}
}
\caption*{\footnotesize
$^\dagger$ \textbf{Task:} D = Detection, S = Segmentation, M = Multi-object tracking. \quad
$^\ddagger$ \textbf{PoV:} E = Ego-centric; F = Fixed (non-CCTV); FC = Fixed CCTV. \quad
$^\ast$ \textbf{Camera count:} UA-DETRAC contains sequences recorded at 24 locations (treated as distinct viewpoints). \quad
$^\diamond$ “--” indicates moving-camera datasets where fixed camera count is not meaningful.
}

\end{table}

To provide a comprehensive view of the datasets commonly used in autonomous driving, surveillance, and aerial vehicle perception research, Table \ref{tab:app-vehicle-datasets-comp} summarizes major moving-camera and fixed-camera benchmarks beyond those directly evaluated in the main paper. These datasets remain influential in computer vision, but differ substantially from our problem setting in terms of viewpoint, task formulation, annotation scope, or geographic context. While they are not used for quantitative comparison due to these mismatches, we include them here to acknowledge their importance and to situate our dataset within the broader ecosystem of vehicle perception resources.

\subsection{Model Training}
\label{app:model-training}

All detectors are trained under a unified protocol with model-specific optimizer settings and augmentation pipelines. For reproducibility, we report all hyperparameters, including learning rates, warmup strategies, augmentation policies, and per-model training schedules, in Table \ref{tab:app-training-hparams-split}.

\begin{table}[ht]
    \centering
    \caption{Training hyperparameter and architectural settings used for all detectors.}
    \resizebox{\linewidth}{!}{
    \begin{tabular}{L{3cm}L{3cm}L{3cm}L{3cm}L{3cm}L{3cm}}
        \hline
        \textbf{Settings}
        & \textbf{YOLOv12-S}
        & \textbf{YOLOv12-X}
        & \textbf{RT-DETRv2-X}
        & \textbf{D-FINE-X}
        & \textbf{RF-DETR-X} \\
        \hline\hline

        \textbf{Batch Size}
        & 16
        & 16
        & 16 
        & 16
        & 16 \\

        \textbf{Epochs}
        & 100
        & 100
        & 100
        & 100
        & 100 \\

        \textbf{Learning Rate}
        & 0.01
        & 0.01
        & $1\times10^{-4}$
        & $2.5\times10^{-4}$
        & $1\times10^{-4}$ \\

        \textbf{Optimizer}
        & AdamW
        & AdamW
        & AdamW
        & AdamW
        & AdamW \\

        \textbf{Weight Decay}
        & $5\times10^{-4}$
        & $5\times10^{-4}$
        & $1\times10^{-4}$
        & $1.25\times10^{-4}$
        & $1\times10^{-4}$ \\

        \textbf{AdamW Betas}
        & (0.937, 0.999)
        & (0.937, 0.999)
        & (0.9, 0.999)
        & (0.9, 0.999)
        & (0.9, 0.999) \\

        \textbf{LR Policy}
        & Cosine
        & Cosine
        & MultiStep
        & MultiStep
        & Step LR \\

        \textbf{Warmup}
        & 3 epochs
        & 3 epochs
        & 2000-iteration linear warmup
        & 500-step linear warmup
        & none \\

        \textbf{Warmup Details}
        & momentum=0.8; bias LR=0.1
        & momentum=0.8; bias LR=0.1
        & momentum untouched; uniform LR ramp
        & no bias/momentum overrides
        & warmup disabled \\

        \textbf{Augmentation Summary}
        & HSV, translate=0.1, scale=0.5, flip=0.5, erase=0.4; no mosaic/mixup
        & HSV, translate=0.1, scale=0.5, flip=0.5, erase=0.4; no mosaic/mixup
        & Photometric, ZoomOut, IoU crop; ops disabled after epoch 151
        & Photometric, ZoomOut, IoU crop, flip, sanitize, resize
        & Flip + multi-scale RandomResize/Crop + normalize \\

        \hline
    \end{tabular}
    }
    \label{tab:app-training-hparams-split}
\end{table}

\subsection{Cross-Domain Evaluations}
\label{app:cross-eval}

\subsubsection{IDD vs.\ \uvhw}
\begin{figure*}[ht!]
        \centering
        \subfloat[AP@50:95 distribution for selected models trained on \uvhw dataset and IDD dataset and evaluated on \uvhw validation split.]{
            \includegraphics[width=1\columnwidth]{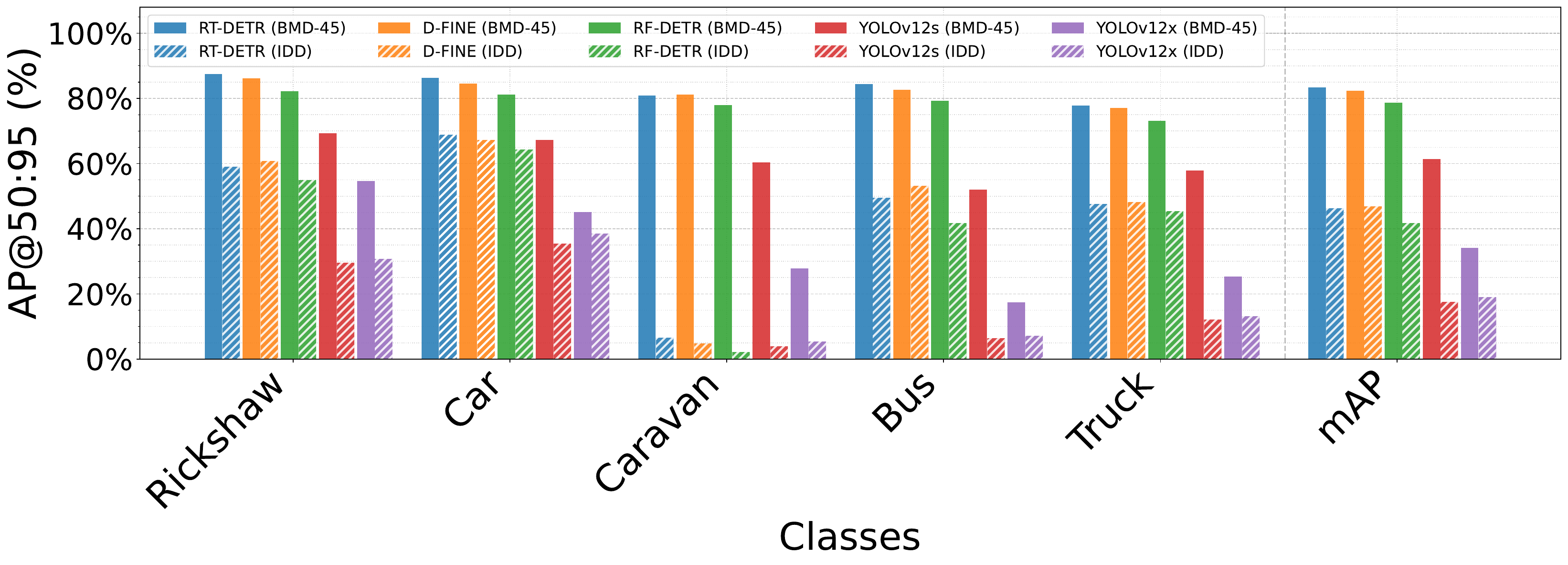}
            \label{fig:model_evaluation_idd_on_ours}            
        }
        \hfill
        \subfloat[AP@50:95 distribution for selected models trained on \uvhw dataset and IDD dataset and evaluated on IDD validation split.~\emph{The Caravan results are not visible as they are near zero due to a low sample count and inconsistent labels}.]{
            \includegraphics[width=1\columnwidth]{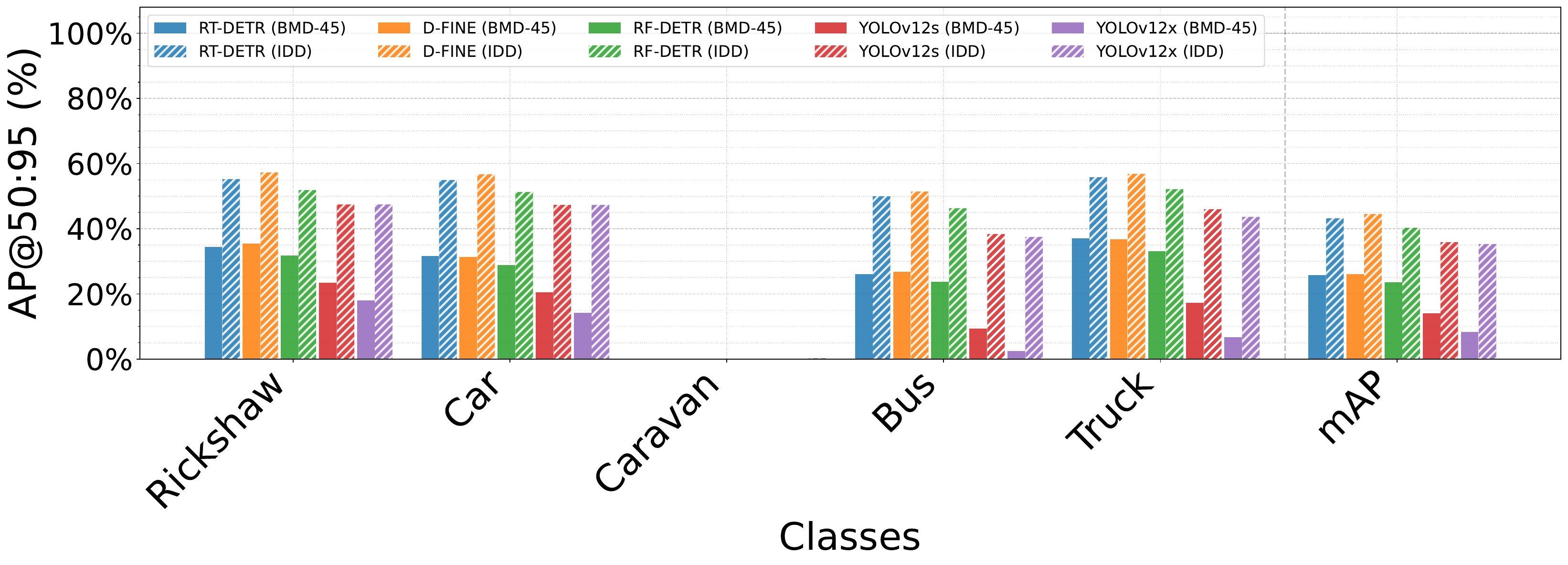}
            \label{fig:model_evaluation_ours_on_idd}
        }
        \caption{Cross-dataset transfer results between IDD and \uvhw.}
        \label{fig:model_evaluation_idd}
\end{figure*}

We analyze cross-dataset transfer between IDD and \uvhw over their shared vehicle categories. When evaluating on the IDD validation split, models trained on IDD consistently outperform those trained on \uvhw, as shown in Figure \ref{fig:model_evaluation_ours_on_idd}. For example, D-FINE trained on IDD achieves 0.444 \mapFiftyNinetyFive on the IDD validation set, whereas the same model trained on \uvhw attains 0.261; similar gaps appear for other models as well.

In the reverse direction, evaluating on the \uvhw validation split, models trained on \uvhw substantially outperform those trained on IDD, as illustrated in Figure \ref{fig:model_evaluation_idd_on_ours}. D-FINE improves from 0.468 (trained on IDD) to 0.823 (trained on \uvhw). YOLOv12 variants show the largest differences, with YOLOv12-S rising from 0.174 to 0.613 and YOLOv12-X from 0.189 to 0.341.

Together, these two-way results (Figure \ref{fig:model_evaluation_idd} and Table~\ref{tab:app_cross_eval_idd_ours}) highlight the strong viewpoint mismatch between ego-centric dashcam footage (IDD) and fixed CCTV imagery (\uvhw), and show that each dataset best supports models trained within its respective domain.

\begin{table}[t]
\centering

\caption{Cross-dataset evaluation -- IDD and \uvhw on adapted classes}
\label{tab:app_cross_eval_idd_ours}
\footnotesize
\setlength{\tabcolsep}{1pt}
\renewcommand{\arraystretch}{1} 
\begin{tabular}{ccc ccc}
\hline
\multicolumn{1}{l}{\textbf{Trained On}} & \multicolumn{1}{l}{\textbf{Evaluated On}} & \textbf{Model} & \textbf{\mapFiftyNinetyFive} & \textbf{\mapSeventyFive} & \textbf{\mapFifty} \\
\hline\hline

\multirow{5}{*}{IDD} & \multirow{5}{*}{IDD} 
& RT-DETRv2 X & 0.431 & 0.449 & 0.603 \\
& & D-FINE X & \textbf{0.444} & \textbf{0.466} & \textbf{0.612} \\
& & RF-DETR X & 0.403 & 0.416 & 0.580 \\
& & YOLOv12 S & 0.358 & 0.376 & 0.513 \\
& & YOLOv12 X & 0.352 & 0.370 & 0.495 \\

\hline

\multirow{5}{*}{\uvhw} & \multirow{5}{*}{IDD} 
& RT-DETRv2 X & 0.258 & 0.279 & 0.376 \\
& & D-FINE X & 0.261 & 0.280 & 0.379 \\
& & RF-DETR X & 0.235 & 0.250 & 0.354 \\
& & YOLOv12 S & 0.141 & 0.155 & 0.208 \\
& & YOLOv12 X & 0.083 & 0.073 & 0.165 \\

\hline\hline

\multirow{5}{*}{IDD} & \multirow{5}{*}{\uvhw} 
& RT-DETRv2 X & 0.463 & 0.514 & 0.585 \\
& & D-FINE X & 0.468 & 0.520 & 0.595 \\
& & RF-DETR X & 0.416 & 0.464 & 0.562 \\
& & YOLOv12 S & 0.174 & 0.190 & 0.260 \\
& & YOLOv12 X & 0.189 & 0.210 & 0.263 \\

\hline

\multirow{5}{*}{\uvhw} & \multirow{5}{*}{\uvhw} 
& RT-DETRv2 X & \textbf{0.833} & \textbf{0.881} & \textbf{0.906} \\
& & D-FINE X & 0.823 & 0.878 & \textbf{0.906} \\
& & RF-DETR X & 0.787 & 0.856 & 0.899 \\
& & YOLOv12 S & 0.613 & 0.673 & 0.748 \\
& & YOLOv12 X & 0.341 & 0.349 & 0.554 \\

\hline
\end{tabular}
\end{table}

\medskip
\subsubsection{UA-DETRAC vs.\ \uvhw}
\begin{figure*}[ht!]
        \centering
        \subfloat[AP@50:95 distribution for selected models trained on \uvhw dataset and UA-DETRAC dataset and evaluated on \uvhw validation split.]{
            \includegraphics[width=1\columnwidth]{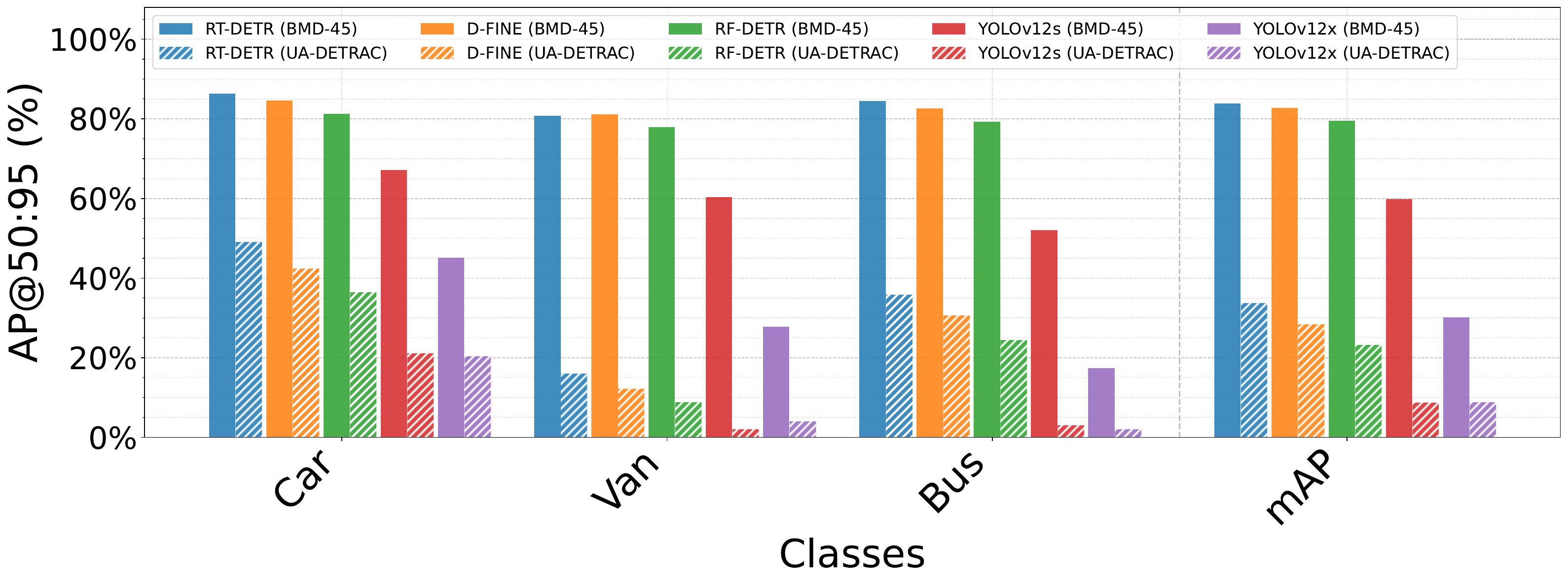}
            \label{fig:model_evaluation_uadetrac_on_ours}            
        }
        \hfill
        \subfloat[AP@50:95 distribution for selected models trained on \uvhw dataset and UA-DETRAC dataset and evaluated on UA-DETRAC validation split.]{
            \includegraphics[width=1\columnwidth]{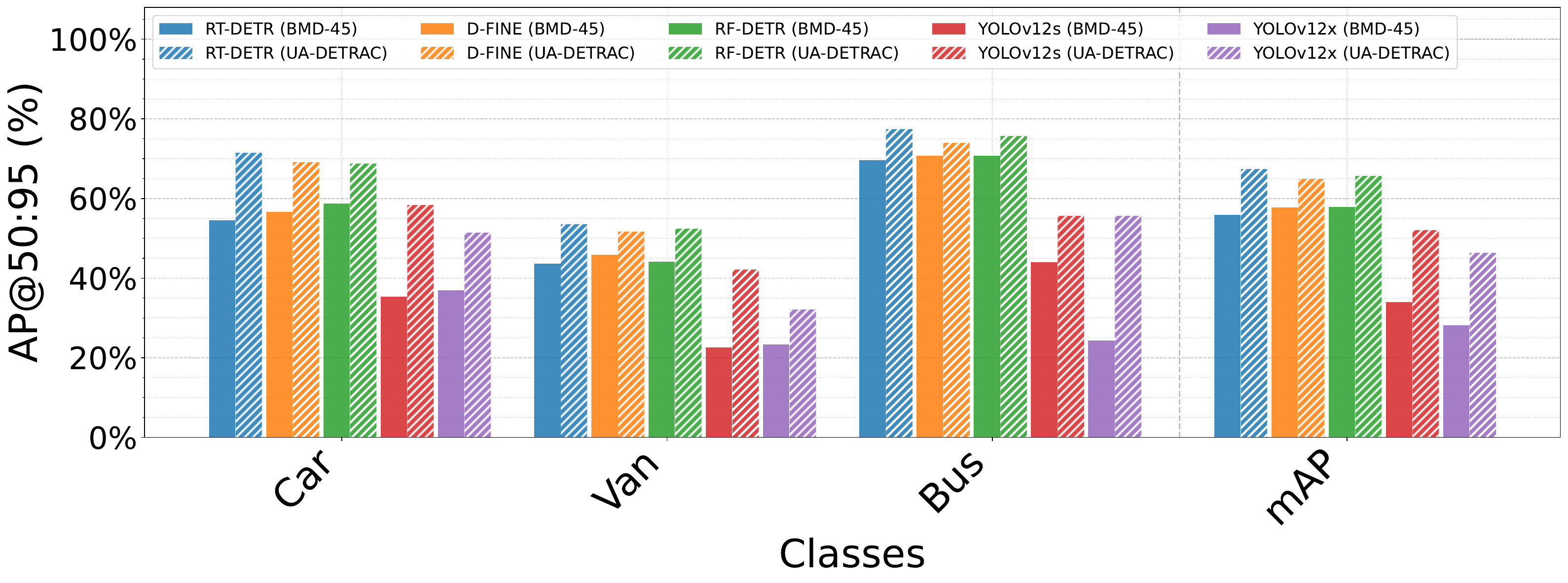}
            \label{fig:model_evaluation_ours_on_uadetrac}
        }
        \caption{Cross-dataset transfer results between UA-DETRAC and \uvhw.}
        \label{fig:model_evaluation_uadetrac}
\end{figure*}

UA-DETRAC exhibits a strong domain mismatch relative to \uvhw, and this asymmetry is clearly reflected in the two-way transfer results (Figure \ref{fig:model_evaluation_uadetrac} and Table~\ref{tab:app_cross_eval_uadetrac_ours}). When evaluated on the \uvhw validation split (Figure \ref{fig:model_evaluation_uadetrac_on_ours}), models trained on UA-DETRAC experience a severe drop in accuracy: 
%D-FINE
RT-DETRv2 X falls from 0.838 \mapFiftyNinetyFive (trained on \uvhw) to 0.336 (trained on UA-DETRAC), with similar declines for other models as well.

In contrast, the reverse direction, evaluating on the UA-DETRAC test split (Figure \ref{fig:model_evaluation_ours_on_uadetrac}), shows a much smaller difference. For 
%D-FINE
RT-DETRv2 X, training on UA-DETRAC yields 0.674 \mapFiftyNinetyFive, only moderately higher than the 0.578 obtained when trained on \uvhw. Other models also follow this pattern.

This asymmetric behavior indicates that models trained on the structured, low-diversity highway scenes of UA-DETRAC generalize poorly to the broader camera viewpoints and richer taxonomy of \uvhw, whereas models trained on \uvhw retain partial transferability back to UA-DETRAC’s narrower domain.

\begin{table}[t]
\centering
\caption{Cross-dataset evaluation -- UA-DETRAC and \uvhw on adapted classes}
\label{tab:app_cross_eval_uadetrac_ours}
\footnotesize
\setlength{\tabcolsep}{1pt}
\renewcommand{\arraystretch}{1} 
\begin{tabular}{ccc ccc}
\hline
\textbf{Trained On}& \textbf{Evaluated On} & \textbf{Model}       
& \textbf{\mapFiftyNinetyFive} & \textbf{\mapSeventyFive} & \textbf{\mapFifty} \\
\hline\hline

\multirow{5}{*}{UA-DETRAC} & \multirow{5}{*}{UA-DETRAC} 
& RT-DETRv2 X & \textbf{0.674} & \textbf{0.778} & \textbf{0.849} \\
& & D-FINE X & 0.649 & 0.754 & 0.821 \\
& & RF-DETR X    & 0.656 & 0.757 & 0.840 \\
& & YOLOv12 S    & 0.520 & 0.604 & 0.718 \\
& & YOLOv12 X    & 0.463 & 0.528 & 0.630 \\

\hline

\multirow{5}{*}{\uvhw} & \multirow{5}{*}{UA-DETRAC} 
& RT-DETRv2 X & 0.559 & 0.681 & 0.798 \\
& & D-FINE X & 0.577 & 0.704 & 0.814 \\
& & RF-DETR X & 0.578 & 0.702 & 0.814 \\
& & YOLOv12 S & 0.340 & 0.388 & 0.540 \\
& & YOLOv12 X & 0.282 & 0.291 & 0.519 \\

\hline\hline

\multirow{5}{*}{UA-DETRAC} & \multirow{5}{*}{\uvhw}
& RT-DETRv2 X & 0.336 & 0.402 & 0.445\\
& & D-FINE X & 0.284 & 0.333 & 0.380 \\
& & RF-DETR X & 0.232 & 0.2730 & 0.308 \\
& & YOLOv12 S & 0.087 & 0.099 & 0.134 \\
& & YOLOv12 X & 0.087 & 0.100 & 0.116\\

\hline

\multirow{5}{*}{\uvhw} & \multirow{5}{*}{\uvhw}
& RT-DETRv2 X & \textbf{0.838} & \textbf{0.881} & \textbf{0.902} \\
& & D-FINE X & 0.828 & 0.877 & 0.900 \\
& & RF-DETR X & 0.795 & 0.859 & 0.893 \\
& & YOLOv12 S & 0.599 & 0.656 & 0.708 \\
& & YOLOv12 X & 0.301 & 0.309 & 0.493 \\

\hline
\end{tabular}
\end{table}

\medskip

\subsubsection{TrafficCAM vs.\ \uvhw}
\begin{figure}[ht!]
        \centering
        \subfloat[AP@50:95 distribution for selected models trained on \uvhw dataset and TrafficCAM dataset and evaluated on \uvhw validation split.]{
            \includegraphics[width=1\columnwidth]{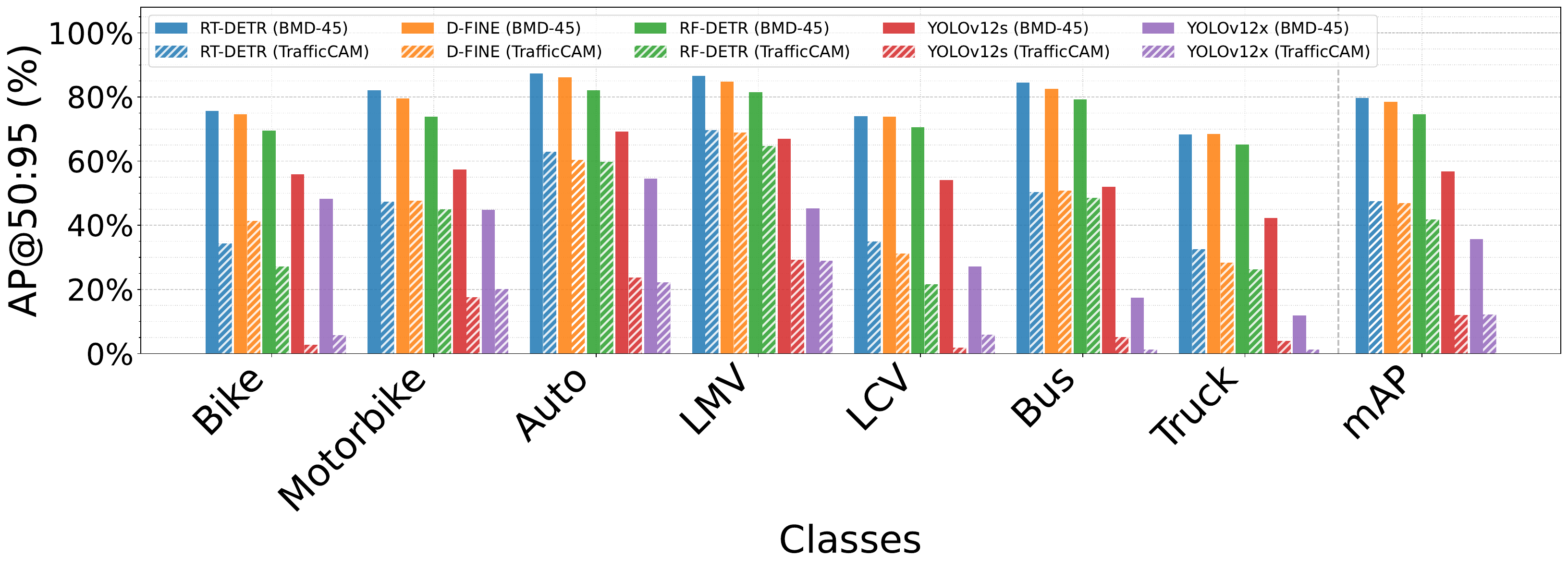}
            \label{fig:model_evaluation_trafficcam_on_ours}            
        }
        \hfill
        \subfloat[AP@50:95 distribution for selected models trained on \uvhw dataset and TrafficCAM dataset and evaluated on TrafficCAM validation split.]{
            \includegraphics[width=1\columnwidth]{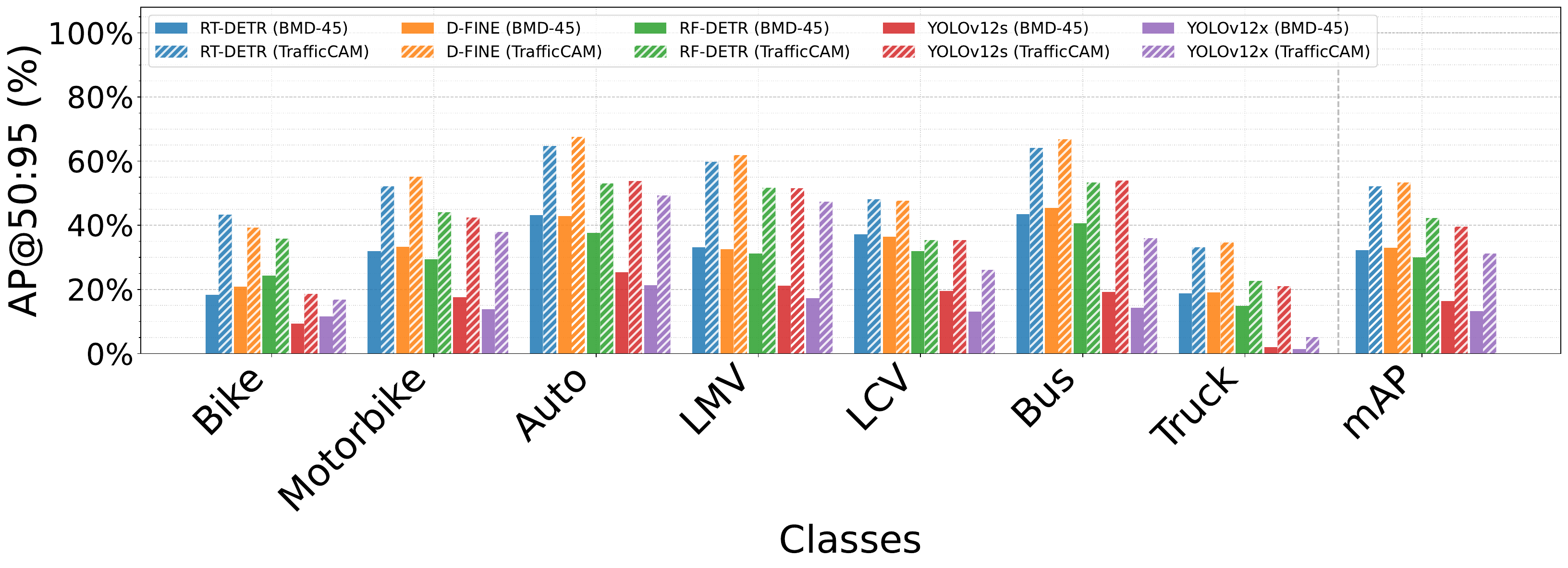}
            \label{fig:model_evaluation_ours_on_trafficcam}
        }
        \caption{Cross-dataset transfer results between TrafficCAM and \uvhw.}
        \label{fig:model_evaluation_trafficcam}
        
\end{figure}

TrafficCAM is the closest to \uvhw in viewpoint and geography, but differences in scale, camera diversity, and annotation policy lead to asymmetric transfer performance (Figure \ref{fig:model_evaluation_trafficcam} and Table~\ref{tab:app_cross_eval_trafficcam_ours}). When evaluated on the \uvhw validation split (Figure \ref{fig:model_evaluation_trafficcam_on_ours}), models trained on TrafficCAM perform noticeably worse than those trained on \uvhw. 
%D-FINE
RT-DETRv2 X drops from 0.798 \mapFiftyNinetyFive (trained on \uvhw) to 0.474 (trained on TrafficCAM); other models follow the same trend.

In the reverse direction, evaluating on the TrafficCAM test split (Figure \ref{fig:model_evaluation_ours_on_trafficcam}), models trained on TrafficCAM outperform those trained on \uvhw, but the margin is smaller.D-FINE achieves 0.532 \mapFiftyNinetyFive when trained on TrafficCAM versus 0.329 when trained on \uvhw; this trend is seen in other models as well.

This asymmetry stems from differences in annotation granularity, label policies, and viewpoint diversity. Together, these factors explain why models trained on \uvhw transfer only partially to TrafficCAM, and likewise why TrafficCAM-trained models underperform on \uvhw, despite the geographic and viewpoint alignment between the two datasets.

\begin{table}[t]
\centering
\caption{Cross-dataset evaluation -- TrafficCAM and \uvhw on adapted classes}
\label{tab:app_cross_eval_trafficcam_ours}
\footnotesize
\setlength{\tabcolsep}{1pt}
\renewcommand{\arraystretch}{1} 
\begin{tabular}{ccc ccc}
\hline
\multicolumn{1}{l}{\textbf{Trained On}} & \multicolumn{1}{l}{\textbf{Evaluated On}} & \textbf{Model} & \textbf{\mapFiftyNinetyFive} & \textbf{\mapSeventyFive} & \textbf{\mapFifty} \\
\hline\hline

\multirow{5}{*}{TrafficCAM} & \multirow{5}{*}{TrafficCAM} 
& RT-DETRv2 X & 0.5214 & 0.561 & \textbf{0.724} \\
& & D-FINE X & \textbf{0.532} & \textbf{0.577} & 0.716 \\
& & RF-DETR X & 0.422 & 0.463 & 0.599 \\
& & YOLOv12 S & 0.394 & 0.429 & 0.566 \\
& & YOLOv12 X & 0.312 & 0.332 & 0.439 \\

\hline

\multirow{5}{*}{\uvhw} & \multirow{5}{*}{TrafficCAM} 
& RT-DETRv2 X & 0.323 & 0.355 & 0.489 \\
& & D-FINE X & 0.329 & 0.364 & 0.496 \\
& & RF-DETR X & 0.300 & 0.323 & 0.470 \\
& & YOLOv12 S & 0.163 & 0.161 & 0.286 \\
& & YOLOv12 X & 0.133 & 0.117 & 0.265 \\

\hline\hline

\multirow{5}{*}{TrafficCAM} & \multirow{5}{*}{\uvhw} 
& RT-DETRv2 X & 0.474 & 0.536 & 0.626 \\
& & D-FINE X & 0.469 & 0.53 & 0.625 \\
& & RF-DETR X & 0.418 & 0.468 & 0.582 \\
& & YOLOv12 S & 0.120 & 0.1290 & 0.192 \\
& & YOLOv12 X & 0.122 & 0.137 & 0.173 \\

\hline

\multirow{5}{*}{\uvhw} & \multirow{5}{*}{\uvhw} 
& RT-DETRv2 X & \textbf{0.798} & \textbf{0.855} & 0.888 \\
& & D-FINE X & 0.786 & 0.849 & \textbf{0.889} \\
& & RF-DETR X & 0.746 & 0.823 & 0.880 \\
& & YOLOv12 S & 0.569 & 0.625 & 0.728 \\
& & YOLOv12 X & 0.356 & 0.363 & 0.586 \\

\hline
\end{tabular}
\end{table}

\begin{figure}[ht]
    \centering
    \includegraphics[width=1\columnwidth]{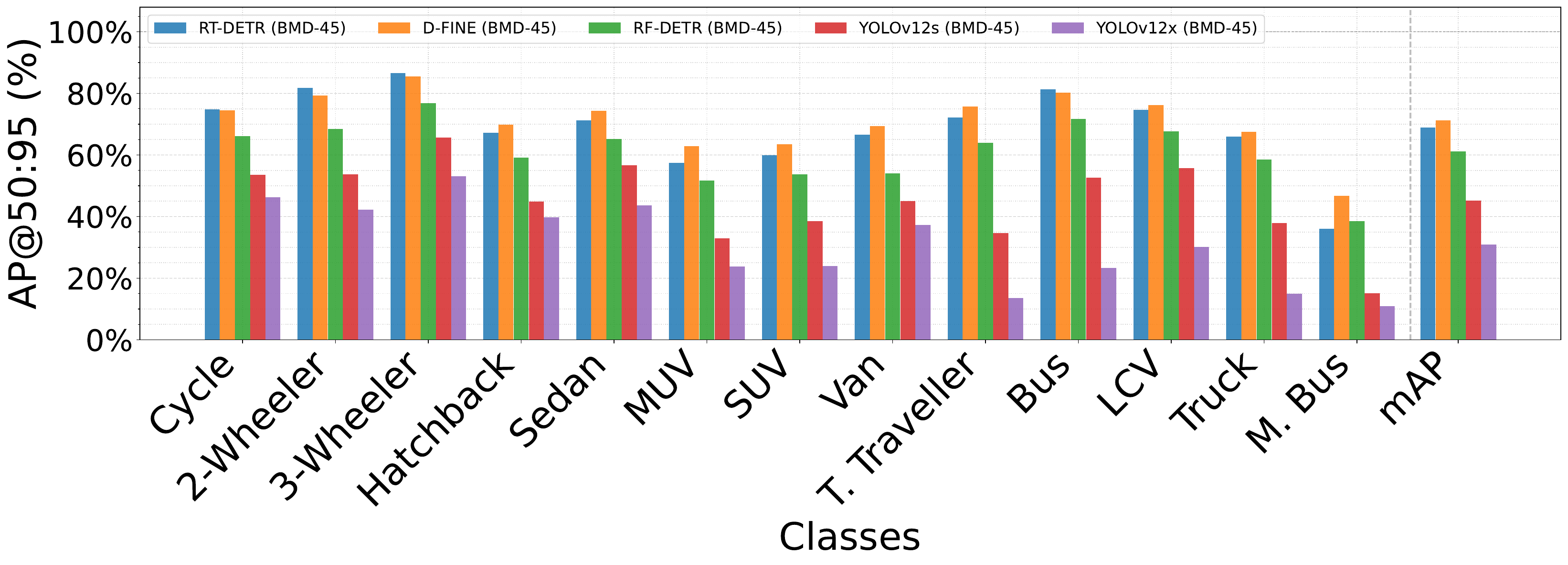}

    \caption{AP@50:95 distribution for selected models trained on \uvhw dataset and evaluated on \uvhw test set.}
    \label{fig:model_evaluation_hackathon_on_hackathon_test}
\end{figure}

\subsection{Evaluation on the Held-Out Test Split}
\label{app:held-out-test}

Following standard practice in benchmark dataset releases (e.g., COCO, Cityscapes), \uvhw publicly provides annotations only for the train and validation splits, while test-set annotations remain private~\cite{lin2014coco, Cordts2015_Cityscapes}. Accordingly, all main paper experiments are reported on the validation split. For completeness, we additionally evaluate models trained on the \uvhw train split on the held-out test split using the private ground truth. As shown in Figure~\ref{fig:model_evaluation_hackathon_on_hackathon_test}, D-FINE achieves 0.712 \mapFiftyNinetyFive, RF-DETR reaches 0.612, and RT-DETR attains 0.689. YOLOv12 models perform comparatively lower, with YOLOv12-S achieving 0.451 and YOLOv12-X reaching 0.31. These results provide a reference for the difficulty of the hidden test set; however, the annotations will not be released to enable future benchmarking protocols.

\subsection{Result Stability Across Random Seeds}
\label{app:random_seeds}

To verify the stability of our reported results, we re-trained RT-DETRv2-X and D-FINE-X on \uvhw with $3$ independent random seeds and evaluated on the \uvhw validation set. RT-DETRv2-X achieved a mAP@$0.50$:$0.95$ of $73.28 \pm 1.04\%$, and D-FINE-X achieved $72.19 \pm 0.06\%$. These are consistent with the values reported in Figure~\ref{fig:model_evaluation_hackathon} ($72.08\%$ and $72.26\%$, respectively), confirming that the observed performance differences across architectures are not artifacts of seed selection. RT-DETRv2-X exhibits slightly higher variance across runs, while D-FINE-X remains stable.

\section{Privacy and Ethical Considerations}
\label{app:ethics}

The \uvhw dataset is derived from fixed-position urban CCTV cameras, which may contain personally identifiable information (PII) such as vehicle license plates, human faces, and on-frame camera metadata. Following established practices in large-scale public computer vision datasets like Street View~\cite{frome2009streetview}, Cityscapes~\cite{cityscapes_scripts}, Waymo Open~\cite{Waymo_FAQ}, and Mapillary~\cite{Mapillary_Privacy}, we apply a structured anonymization protocol before public release. All privacy-preserving transformations were performed \emph{after} the completion of annotation, ensuring that volunteers only interacted with unblurred imagery during labeling to maintain accuracy for small objects and fine-grained vehicle classes. No identifiable annotator information is released with the dataset.

\subsection{License Plates}
License plates are detected using a YOLO-based one-stage detector trained for road-traffic imagery. Detected regions are blurred with a Gaussian kernel whose size is proportional to the bounding-box dimensions, ensuring complete removal of alphanumeric content across varying viewpoints and resolutions. Multi-scale inference is applied to reliably identify distant or low-resolution plates. This approach maintains local visual appearance while eliminating identifiable text, consistent with redaction strategies widely used in traffic and street-view data~\cite{Waymo_FAQ,Mapillary_Privacy}.

\subsection{Faces}
Faces are detected using a modern SCRFD-based face detector~\cite{scrfd2021}. To improve robustness under CCTV-specific conditions, like directional lighting, low contrast, and varied skin tones, we apply lightweight pre-processing steps including white-balance correction, contrast-limited adaptive histogram equalization (CLAHE), gamma adjustment, and unsharp masking prior to detection. Multi-scale and tiled inference is used to capture small or partially occluded faces. Each detected region is expanded by $20\%$ and blurred with an adaptive Gaussian kernel, obfuscating identity while preserving overall scene structure. Prior work suggests that such redaction substantially reduces re-identification risk while maintaining utility for downstream scene-understanding tasks~\cite{dietlmeier2021facesreid}.

\subsection{On-frame Camera Overlays}
CCTV feeds occasionally contain textual overlays such as camera IDs, timestamps, and location descriptors. To remove these, we apply an OCR-based redaction pipeline using PP-OCRv3~\cite{ppocrv3}. Predefined regions (e.g., corners and header bands) are scanned for text; detected polygons are expanded by a small margin to ensure complete coverage. Identified regions are then removed through fast-marching inpainting~\cite{telea2004} in OpenCV~\cite{opencv_bradski2000}, which reconstructs background content using adjacent pixel information. This removes contextual identifiers without introducing large uniform patches that could bias model training.

\subsection{Scope and Limitations}
Our anonymization process targets the dominant PII categories present in fixed-view CCTV imagery. While no automated pipeline can guarantee absolute removal of all identifying content, the combination of multi-stage detection, adaptive blurring, and inpainting yields privacy protection comparable to or exceeding that of existing public benchmarks. The final released dataset contains only redacted frames, with all original unblurred images retained offline solely for annotation and evaluation conducted by the authors.

%% file: main.bib
@String(CVPR= {IEEE Conf. Comput. Vis. Pattern Recog.})

@String(ICCV= {Int. Conf. Comput. Vis.})

@String(ECCV= {Eur. Conf. Comput. Vis.})

@String(ICPR = {Int. Conf. Pattern Recog.})

@String(AAAI = {AAAI})

@String(CVPR  = {CVPR})

@String(ICCV  = {ICCV})

@String(ECCV  = {ECCV})

@String(ICPR  = {ICPR})

@INPROCEEDINGS{Tang2019_CityFlow,
author={Tang, Zheng and Naphade, Milind and Liu, Ming-Yu and Yang, Xiaodong and Birchfield, Stan and Wang, Shuo and Kumar, Ratnesh and Anastasiu, David and Hwang, Jenq-Neng},
booktitle={2019 IEEE/CVF Conference on Computer Vision and Pattern Recognition (CVPR)}, 
title={CityFlow: A City-Scale Benchmark for Multi-Target Multi-Camera Vehicle Tracking and Re-Identification}, 
year={2019},
volume={},
number={},
pages={8789-8798},
doi={10.1109/CVPR.2019.00900}
}

@INPROCEEDINGS{Geiger2012_KITTI,
author={Geiger, Andreas and Lenz, Philip and Urtasun, Raquel},
booktitle={2012 IEEE Conference on Computer Vision and Pattern Recognition}, 
title={Are we ready for autonomous driving? The KITTI vision benchmark suite}, 
year={2012},
volume={},
number={},
pages={3354-3361},
doi={10.1109/CVPR.2012.6248074}
}

@inproceedings{Yu2020_BDD100K,
title = {BDD100K: A Diverse Driving Dataset for Heterogeneous Multitask Learning},
url = {http://dx.doi.org/10.1109/CVPR42600.2020.00271},
DOI = {10.1109/cvpr42600.2020.00271},
booktitle = {2020 IEEE/CVF Conference on Computer Vision and Pattern Recognition (CVPR)},
publisher = {IEEE},
author = {Yu,  Fisher and Chen,  Haofeng and Wang,  Xin and Xian,  Wenqi and Chen,  Yingying and Liu,  Fangchen and Madhavan,  Vashisht and Darrell,  Trevor},
year = {2020},
month = jun,
pages = {2633–2642}
}

@article{Wen2020_UADETRAC,
title = {UA-DETRAC: A new benchmark and protocol for multi-object detection and tracking},
journal = {Computer Vision and Image Understanding},
volume = {193},
pages = {102907},
year = {2020},
issn = {1077-3142},
doi = {https://doi.org/10.1016/j.cviu.2020.102907},
url = {https://www.sciencedirect.com/science/article/pii/S1077314220300035},
author = {Longyin Wen and Dawei Du and Zhaowei Cai and Zhen Lei and Ming-Ching Chang and Honggang Qi and Jongwoo Lim and Ming-Hsuan Yang and Siwei Lyu},
}

@ARTICLE{Deng2025_TrafficCAM,
author={Deng, Zhongying and Cheng, Yanqi and Liu, Lihao and Wang, Shujun and Ke, Rihuan and Schönlieb, Carola-Bibiane and Aviles-Rivero, Angelica I.},
journal={IEEE Transactions on Intelligent Transportation Systems}, 
title={TrafficCAM: A Versatile Dataset for Traffic Flow Segmentation}, 
year={2025},
volume={26},
number={2},
pages={2747-2759},
doi={10.1109/TITS.2024.3510551}
}

@INPROCEEDINGS{Zhang2017_CityCam,
author={Zhang, Shanghang and Wu, Guanhang and Costeira, João P. and Moura, José M. F.},
booktitle={2017 IEEE Conference on Computer Vision and Pattern Recognition (CVPR)}, 
title={Understanding Traffic Density from Large-Scale Web Camera Data}, 
year={2017},
volume={},
number={},
pages={4264-4273},
doi={10.1109/CVPR.2017.454}
}

@article{Geiger2013_KITTI,
author = {Geiger, A and Lenz, P and Stiller, C and Urtasun, R},
title = {Vision meets robotics: The KITTI dataset},
year = {2013},
issue_date = {September 2013},
publisher = {Sage Publications, Inc.},
address = {USA},
volume = {32},
number = {11},
issn = {0278-3649},
url = {https://doi.org/10.1177/0278364913491297},
doi = {10.1177/0278364913491297},
journal = {Int. J. Rob. Res.},
month = sep,
pages = {1231–1237},
numpages = {7}
}

@INPROCEEDINGS {Cordts2015_Cityscapes,
author = { Cordts, Marius and Omran, Mohamed and Ramos, Sebastian and Rehfeld, Timo and Enzweiler, Markus and Benenson, Rodrigo and Franke, Uwe and Roth, Stefan and Schiele, Bernt },
booktitle = { 2016 IEEE Conference on Computer Vision and Pattern Recognition (CVPR) },
title = {{ The Cityscapes Dataset for Semantic Urban Scene Understanding }},
year = {2016},
volume = {},
ISSN = {1063-6919},
pages = {3213-3223},
doi = {10.1109/CVPR.2016.350},
url = {https://doi.ieeecomputersociety.org/10.1109/CVPR.2016.350},
publisher = {IEEE Computer Society},
address = {Los Alamitos, CA, USA},
month =Jun}

@INPROCEEDINGS {Caesar2020_nuScenes,
author = { Caesar, Holger and Bankiti, Varun and Lang, Alex H. and Vora, Sourabh and Liong, Venice Erin and Xu, Qiang and Krishnan, Anush and Pan, Yu and Baldan, Giancarlo and Beijbom, Oscar },
booktitle = { 2020 IEEE/CVF Conference on Computer Vision and Pattern Recognition (CVPR) },
title = {{ nuScenes: A Multimodal Dataset for Autonomous Driving }},
year = {2020},
volume = {},
ISSN = {},
pages = {11618-11628},
doi = {10.1109/CVPR42600.2020.01164},
url = {https://doi.ieeecomputersociety.org/10.1109/CVPR42600.2020.01164},
publisher = {IEEE Computer Society},
address = {Los Alamitos, CA, USA},
month =Jun
}

@INPROCEEDINGS{Varma2019_IDD,
author={Varma, Girish and Subramanian, Anbumani and Namboodiri, Anoop and Chandraker, Manmohan and Jawahar, C.V.},
booktitle={2019 IEEE Winter Conference on Applications of Computer Vision (WACV)}, 
title={IDD: A Dataset for Exploring Problems of Autonomous Navigation in Unconstrained Environments}, 
year={2019},
volume={},
number={},
pages={1743-1751},
doi={10.1109/WACV.2019.00190}
}

@article{pascalVOC2010everingham,
author = {Everingham, Mark and Van Gool, Luc and Williams, Christopher and Winn, John and Zisserman, Andrew},
year = {2010},
month = {06},
pages = {303-338},
title = {The Pascal Visual Object Classes (VOC) challenge},
volume = {88},
journal = {International Journal of Computer Vision},
doi = {10.1007/s11263-009-0275-4}
}

@InProceedings{lin2014coco,
author="Lin, Tsung-Yi
and Maire, Michael
and Belongie, Serge
and Hays, James
and Perona, Pietro
and Ramanan, Deva
and Doll{\'a}r, Piotr
and Zitnick, C. Lawrence",
editor="Fleet, David
and Pajdla, Tomas
and Schiele, Bernt
and Tuytelaars, Tinne",
title="Microsoft COCO: Common Objects in Context",
booktitle="Computer Vision -- ECCV 2014",
year="2014",
publisher="Springer International Publishing",
address="Cham",
pages="740--755",
isbn="978-3-319-10602-1"
}

@inproceedings{du2018uavdt,
author = {Du, Dawei and Qi, Yuankai and Yu, Hongyang and Yang, Yifan and Duan, Kaiwen and Li, Guorong and Zhang, Weigang and Huang, Qingming and Tian, Qi},
title = {The Unmanned Aerial Vehicle Benchmark: Object Detection and Tracking},
year = {2018},
isbn = {978-3-030-01248-9},
publisher = {Springer-Verlag},
address = {Berlin, Heidelberg},
url = {https://doi.org/10.1007/978-3-030-01249-6_23},
doi = {10.1007/978-3-030-01249-6_23},
booktitle = {Computer Vision – ECCV 2018: 15th European Conference, Munich, Germany, September 8-14, 2018, Proceedings, Part X},
pages = {375–391},
numpages = {17},
location = {Munich, Germany}
}

@misc{jocher2023yolov8,
  author = {Glenn Jocher and Ayush Chaurasia and Jing Qiu},
  title = {Ultralytics YOLOv8},
  version = {8.0.0},
  year = {2023},
  url = {https://github.com/ultralytics/ultralytics},
  orcid = {0000-0001-5950-6979, 0000-0002-7603-6750, 0000-0003-3783-7069},
  license = {AGPL-3.0}
}

@misc{jocher2024yolov11,
  author = {Glenn Jocher and Jing Qiu},
  title = {Ultralytics YOLO11},
  version = {11.0.0},
  year = {2024},
  url = {https://github.com/ultralytics/ultralytics},
  orcid = {0000-0001-5950-6979, 0000-0003-3783-7069},
  license = {AGPL-3.0}
}

@misc{xu2022damo,
  doi = {10.48550/ARXIV.2211.15444},
  url = {https://arxiv.org/abs/2211.15444},
  author = {Xu,  Xianzhe and Jiang,  Yiqi and Chen,  Weihua and Huang,  Yilun and Zhang,  Yuan and Sun,  Xiuyu},
  title = {DAMO-YOLO : A Report on Real-Time Object Detection Design},
  publisher = {arXiv},
  year = {2022},
  copyright = {arXiv.org perpetual,  non-exclusive license}
}

@misc{Lv2024_RTDETRv2,
title={RT-DETRv2: Improved Baseline with Bag-of-Freebies for Real-Time Detection Transformer}, 
author={Wenyu Lv and Yian Zhao and Qinyao Chang and Kui Huang and Guanzhong Wang and Yi Liu},
year={2024},
eprint={2407.17140},
archivePrefix={arXiv},
primaryClass={cs.CV},
url={https://arxiv.org/abs/2407.17140}, 
}

@inproceedings{peng2025dfine,
    title={D-{FINE}: Redefine Regression Task of {DETR}s as Fine-grained Distribution Refinement},
    author={Yansong Peng and Hebei Li and Peixi Wu and Yueyi Zhang and Xiaoyan Sun and Feng Wu},
    booktitle={The Thirteenth International Conference on Learning Representations},
    year={2025},
    url={https://openreview.net/forum?id=MFZjrTFE7h}
}

@misc{Robinson2025_RFDETR,
    title={RF-DETR: Neural Architecture Search for Real-Time Detection Transformers}, 
    author={Isaac Robinson and Peter Robicheaux and Matvei Popov and Deva Ramanan and Neehar Peri},
    year={2025},
    eprint={2511.09554},
    archivePrefix={arXiv},
    primaryClass={cs.CV},
    url={https://arxiv.org/abs/2511.09554}, 
}

@inproceedings{Liu2024_DINO,
author = {Liu, Shilong and Zeng, Zhaoyang and Ren, Tianhe and Li, Feng and Zhang, Hao and Yang, Jie and Jiang, Qing and Li, Chunyuan and Yang, Jianwei and Su, Hang and Zhu, Jun and Zhang, Lei},
title = {Grounding DINO: Marrying DINO with\&nbsp;Grounded Pre-training for\&nbsp;Open-Set Object Detection},
year = {2024},
isbn = {978-3-031-72969-0},
publisher = {Springer-Verlag},
address = {Berlin, Heidelberg},
url = {https://doi.org/10.1007/978-3-031-72970-6_3},
doi = {10.1007/978-3-031-72970-6_3},
booktitle = {Computer Vision – ECCV 2024: 18th European Conference, Milan, Italy, September 29–October 4, 2024, Proceedings, Part XLVII},
pages = {38–55},
numpages = {18},
location = {Milan, Italy}
}

@Article{Peppa2021,
AUTHOR = {Peppa, Maria V. and Komar, Tom and Xiao, Wen and James, Phil and Robson, Craig and Xing, Jin and Barr, Stuart},
TITLE = {Towards an End-to-End Framework of CCTV-Based Urban Traffic Volume Detection and Prediction},
JOURNAL = {Sensors},
VOLUME = {21},
YEAR = {2021},
NUMBER = {2},
ARTICLE-NUMBER = {629},
URL = {https://www.mdpi.com/1424-8220/21/2/629},
PubMedID = {33477471},
ISSN = {1424-8220},
DOI = {10.3390/s21020629}
}

@INPROCEEDINGS{Chen2018,
author={Chen, Yuhua and Li, Wen and Sakaridis, Christos and Dai, Dengxin and Van Gool, Luc},
booktitle={2018 IEEE/CVF Conference on Computer Vision and Pattern Recognition}, 
title={Domain Adaptive Faster R-CNN for Object Detection in the Wild}, 
year={2018},
volume={},
number={},
pages={3339-3348},
doi={10.1109/CVPR.2018.00352}
}

@article{Zhou2025,
author = {Zhou, Wei and Yang, Li and Zhao, Lei and Zhang, Runyu and Cui, Yifan and Huang, Hongpu and Qie, Kun and Wang, Chen},
title = {Vision Technologies with Applications in Traffic Surveillance Systems: A Holistic Survey},
year = {2025},
issue_date = {February 2026},
publisher = {Association for Computing Machinery},
address = {New York, NY, USA},
volume = {58},
number = {3},
issn = {0360-0300},
url = {https://doi.org/10.1145/3760525},
doi = {10.1145/3760525},
journal = {ACM Comput. Surv.},
month = sep,
articleno = {58},
numpages = {47}
}

@Article{Myagmar2023,
AUTHOR = {Myagmar-Ochir, Yanjinlkham and Kim, Wooseong},
TITLE = {A Survey of Video Surveillance Systems in Smart City},
JOURNAL = {Electronics},
VOLUME = {12},
YEAR = {2023},
NUMBER = {17},
ARTICLE-NUMBER = {3567},
URL = {https://www.mdpi.com/2079-9292/12/17/3567},
ISSN = {2079-9292},
DOI = {10.3390/electronics12173567}
}

@Article{Pi2022,
author={Pi, Yalong
and Duffield, Nick
and Behzadan, Amir H.
and Lomax, Tim},
title={Visual recognition for urban traffic data retrieval and analysis in major events using convolutional neural networks},
journal={Computational Urban Science},
year={2022},
month={Jan},
day={06},
volume={2},
number={1},
pages={2},
issn={2730-6852},
doi={10.1007/s43762-021-00031-w},
url={https://doi.org/10.1007/s43762-021-00031-w}
}

@INPROCEEDINGS{lyu2023box,
  author={Lyu, Mengyao and Zhou, Jundong and Chen, Hui and Huang, Yijie and Yu, Dongdong and Li, Yaqian and Guo, Yandong and Guo, Yuchen and Xiang, Liuyu and Ding, Guiguang},
  booktitle={2023 IEEE/CVF Conference on Computer Vision and Pattern Recognition (CVPR)}, 
  title={Box-Level Active Detection}, 
  year={2023},
  volume={},
  number={},
  pages={23766-23775},
  doi={10.1109/CVPR52729.2023.02276}}

@INPROCEEDINGS{wang2018ssm,
  author={Wang, Keze and Yan, Xiaopeng and Zhang, Dongyu and Zhang, Lei and Lin, Liang},
  booktitle={2018 IEEE/CVF Conference on Computer Vision and Pattern Recognition}, 
  title={Towards Human-Machine Cooperation: Self-Supervised Sample Mining for Object Detection}, 
  year={2018},
  volume={},
  number={},
  pages={1605-1613},
  doi={10.1109/CVPR.2018.00173}}

@Article{Yu2023,
author={Yu, Fudan
and Yan, Huan
and Chen, Rui
and Zhang, Guozhen
and Liu, Yu
and Chen, Meng
and Li, Yong},
title={City-scale Vehicle Trajectory Data from Traffic Camera Videos},
journal={Scientific Data},
year={2023},
month={Oct},
day={17},
volume={10},
number={1},
pages={711},
issn={2052-4463},
doi={10.1038/s41597-023-02589-y},
url={https://doi.org/10.1038/s41597-023-02589-y}
}

@misc{Huang2025,
title={What Demands Attention in Urban Street Scenes? From Scene Understanding towards Road Safety: A Survey of Vision-driven Datasets and Studies}, 
author={Yaoqi Huang and Julie Stephany Berrio and Mao Shan and Stewart Worrall},
year={2025},
eprint={2507.06513},
archivePrefix={arXiv},
primaryClass={cs.CV},
url={https://arxiv.org/abs/2507.06513}, 
}

@article{Liu2023,
title = {Deep transfer learning for intelligent vehicle perception: A survey},
journal = {Green Energy and Intelligent Transportation},
volume = {2},
number = {5},
pages = {100125},
year = {2023},
issn = {2773-1537},
doi = {https://doi.org/10.1016/j.geits.2023.100125},
url = {https://www.sciencedirect.com/science/article/pii/S2773153723000610},
author = {Xinyu Liu and Jinlong Li and Jin Ma and Huiming Sun and Zhigang Xu and Tianyun Zhang and Hongkai Yu},
}

@article{Thatipelli2025,
title = {Egocentric and exocentric methods: A short survey},
journal = {Computer Vision and Image Understanding},
volume = {257},
pages = {104371},
year = {2025},
issn = {1077-3142},
doi = {https://doi.org/10.1016/j.cviu.2025.104371},
url = {https://www.sciencedirect.com/science/article/pii/S1077314225000943},
author = {Anirudh Thatipelli and Shao-Yuan Lo and Amit K. Roy-Chowdhury},
}

@Misc{Klein2006,
author={Klein, Lawrence A.
and Mills, Milton K.
and Gibson, David},
title={Traffic Detector Handbook: Third edition. Volume II},
year={2006},
month={Oct},
day={01},
note={Tech Report -- FHWA-HRT-06-139},
url={https://rosap.ntl.bts.gov/view//936},
language={English}
}

@INPROCEEDINGS{Xiang2013,
author={Xiang, Yu and Savarese, Silvio},
booktitle={2013 IEEE International Conference on Computer Vision Workshops}, 
title={Object Detection by 3D Aspectlets and Occlusion Reasoning}, 
year={2013},
volume={},
number={},
pages={530-537},
doi={10.1109/ICCVW.2013.75}
}

@inproceedings{Tian2025_YOLOv12,
title={{YOLO}v12: Attention-Centric Real-Time Object Detectors},
author={Yunjie Tian and Qixiang Ye and David Doermann},
booktitle={The Thirty-ninth Annual Conference on Neural Information Processing Systems},
year={2025},
url={https://openreview.net/forum?id=gCvByDI4FN}
}

@INPROCEEDINGS {Sun2020_Waymo,
author = { Sun, Pei and Kretzschmar, Henrik and Dotiwalla, Xerxes and Chouard, Aurelien and Patnaik, Vijaysai and Tsui, Paul and Guo, James and Zhou, Yin and Chai, Yuning and Caine, Benjamin and Vasudevan, Vijay and Han, Wei and Ngiam, Jiquan and Zhao, Hang and Timofeev, Aleksei and Ettinger, Scott and Krivokon, Maxim and Gao, Amy and Joshi, Aditya and Zhang, Yu and Shlens, Jonathon and Chen, Zhifeng and Anguelov, Dragomir },
booktitle = { 2020 IEEE/CVF Conference on Computer Vision and Pattern Recognition (CVPR) },
title = {{ Scalability in Perception for Autonomous Driving: Waymo Open Dataset }},
year = {2020},
volume = {},
ISSN = {},
pages = {2443-2451},
doi = {10.1109/CVPR42600.2020.00252},
url = {https://doi.ieeecomputersociety.org/10.1109/CVPR42600.2020.00252},
publisher = {IEEE Computer Society},
address = {Los Alamitos, CA, USA},
month =Jun
}

@article{Zhou2019,
author = {Zhou, Yao and Ying, Lei and He, Jingrui},
title = {Multi-task Crowdsourcing via an Optimization Framework},
year = {2019},
issue_date = {June 2019},
publisher = {Association for Computing Machinery},
address = {New York, NY, USA},
volume = {13},
number = {3},
issn = {1556-4681},
url = {https://doi.org/10.1145/3310227},
doi = {10.1145/3310227},
journal = {ACM Trans. Knowl. Discov. Data},
month = may,
articleno = {27},
numpages = {26}
}

@article{Sheshadri2013,
title        = {SQUARE: A Benchmark for Research on Computing Crowd Consensus},
volume       = {1},
url          = {https://ojs.aaai.org/index.php/HCOMP/article/view/13088},
doi          = {10.1609/hcomp.v1i1.13088},
number       = {1},
journal      = {Proceedings of the AAAI Conference on Human Computation and Crowdsourcing},
author       = {Sheshadri, Aashish and Lease, Matthew},
year         = {2013},
month        = {Nov.},
pages        = {156-164}
}

@inproceedings{frome2009streetview,
  author    = {Frome, Andrea and Cheung, George S. and Abdulkader, Ahmed and Zennaro, Marco and Wu, Bo and Bissacco, Alessandro and Adam, Hartmut and Neven, Hartmut and Vincent, Luc},
  title     = {Large-scale privacy protection in Google Street View},
  booktitle = {Proceedings of the IEEE International Conference on Computer Vision (ICCV)},
  year      = {2009}
}

@misc{Waymo_FAQ,
  author       = {{Waymo Research}},
  title        = {Waymo Open Dataset FAQ: ``What are you doing to ensure the privacy of people in the images?''},
  howpublished = {\url{https://waymo.com/open/faq/}},
  note         = {Accessed November 2025}
}

@misc{Mapillary_Privacy,
  author       = {{Mapillary}},
  title        = {Privacy Policy and Help Center: automatic blurring of faces and license plates},
  howpublished = {\url{https://www.mapillary.com/privacy}},
  note         = {Accessed November 2025. See also \url{https://help.mapillary.com/hc/en-us/articles/115001663705-Blurring-images-on-Mapillary}}
}

@misc{cityscapes_scripts,
  author       = {Cordts, Marius},
  title        = {Prepare Cityscapes dataset (notes on blurred images)},
  howpublished = {\url{https://github.com/mcordts/cityscapesScripts}},
  note         = {Cityscapes Scripts, GitHub repository, accessed November 2025}
}

@INPROCEEDINGS{visdrone,
  author={Du, Dawei and Zhu, Pengfei and Wen, Longyin and Bian, Xiao and Lin, Haibin and Hu, Qinghua and Peng, Tao and Zheng, Jiayu and Wang, Xinyao and Zhang, Yue and Bo, Liefeng and Shi, Hailin and Zhu, Rui and Kumar, Aashish and Li, Aijin and Zinollayev, Almaz and Askergaliyev, Anuar and Schumann, Arne and Mao, Binjie and Lee, Byeongwon and Liu, Chang and Chen, Changrui and Pan, Chunhong and Huo, Chunlei and Yu, Da and Cong, DeChun and Zeng, Dening and Pailla, Dheeraj Reddy and Li, Di and Wang, Dong and Cho, Donghyeon and Zhang, Dongyu and Bai, Furui and Jose, George and Gao, Guangyu and Liu, Guizhong and Xiong, Haitao and Qi, Hao and Wang, Haoran and Qiu, Heqian and Li, HongLiang and Lu, Huchuan and Kim, Ildoo and Kim, Jaekyum and Shen, Jane and Lee, Jihoon and Ge, Jing and Xu, Jingjing and Zhou, Jingkai and Meier, Jonas and Choi, Jun Won and Hu, Junhao and Zhang, Junyi and Huang, Junying and Huang, Kaiqi and Wang, Keyang and Sommer, Lars and Jin, Lei and Zhang, Lei and Huang, Lianghua and Sun, Lin and Steinmann, Lucas and Jia, Meixia and Xu, Nuo and Zhang, Pengyi and Chen, Qiang and Lv, Qingxuan and Liu, Qiong and Cheng, Qishang and Chennamsetty, Sai Saketh and Chen, Shuhao and Wei, Shuo and Kruthiventi, Srinivas S S and Hong, Sungeun and Kang, Sungil and Wu, Tong and Feng, Tuo and Kollerathu, Varghese Alex and Li, Wanqi and Dai, Wei and Qin, Weida and Wang, Weiyang and Wang, Xiaorui and Chen, Xiaoyu and Chen, Xin and Sun, Xin and Zhang, Xin and Zhao, Xin and Zhang, Xindi and Zhang, Xinyu and Chen, Xuankun and Wei, Xudong and Zhang, Xuzhang and Li, Yanchao and Chen, Yifu and Toh, Yu Heng and Zhang, Yu and Zhu, Yu and Zhong, Yunxin and Wang, Zexin and Wang, Zhikang and Song, Zichen and Liu, Ziming},
  booktitle={2019 IEEE/CVF International Conference on Computer Vision Workshop (ICCVW)}, 
  title={VisDrone-DET2019: The Vision Meets Drone Object Detection in Image Challenge Results}, 
  year={2019},
  volume={},
  number={},
  pages={213-226},
  doi={10.1109/ICCVW.2019.00030}
}

@article{scrfd2021,
  author  = {Guo, Jia and Deng, Jiankang and Lattas, Andreas and Zafeiriou, Stefanos},
  title   = {Sample and Computation Redistribution for Efficient Face Detection (SCRFD)},
  journal = {arXiv preprint arXiv:2105.04714},
  year    = {2021}
}

@INPROCEEDINGS {dietlmeier2021facesreid,
author = { Dietlmeier, Julia and Antony, Joseph and McGuinness, Kevin and O'Connor, Noel E. },
booktitle = { 2020 25th International Conference on Pattern Recognition (ICPR) },
title = {{ How important are faces for person re-identification? }},
year = {2021},
volume = {},
ISSN = {1051-4651},
pages = {6912-6919},
doi = {10.1109/ICPR48806.2021.9412340},
url = {https://doi.ieeecomputersociety.org/10.1109/ICPR48806.2021.9412340},
publisher = {IEEE Computer Society},
address = {Los Alamitos, CA, USA},
month =Jan}

@article{ppocrv3,
  author  = {Li, Chen and Liu, Wei and Guo, Ruoyu and Yin, Xiaohui and Jiang, Kai and Du, Yuning and Du, Yuning and Zhu, Liang and Lai, Bo and Hu, Xin and Yu, Dian and Ma, Yi},
  title   = {PP-OCRv3: More Attempts for the Improvement of Ultra Lightweight OCR System},
  journal = {arXiv preprint arXiv:2206.03001},
  year    = {2022}
}

@article{telea2004,
  author  = {Telea, Alexandru},
  title   = {An Image Inpainting Technique Based on the Fast Marching Method},
  journal = {Journal of Graphics Tools},
  volume  = {9},
  number  = {1},
  pages   = {23--34},
  year    = {2004}
}

@article{opencv_bradski2000,
  author  = {Bradski, Gary},
  title   = {The OpenCV Library},
  journal = {Dr. Dobb's Journal of Software Tools},
  year    = {2000},
  note    = {\url{https://docs.opencv.org/}}
}

@article{he2024enhancinguadetrac,
author = {He, Jinpeng and Chen, Huaixin and Liu, Biyuan and Luo, Sijie and Liu, Jie},
year = {2024},
month = {08},
pages = {},
title = {Enhancing YOLO for occluded vehicle detection with grouped orthogonal attention and dense object repulsion},
volume = {14},
journal = {Scientific Reports},
doi = {10.1038/s41598-024-70695-x}
}

@misc{TomTom2025,
author       = "TomTom",
title        = "Annual TomTom Traffic Index: Unveiling data-driven insights from over 450 billion miles driven in 2024",
howpublished = "https://www.tomtom.com/traffic-index/ranking/",
month        = "Jan",
year         = "2025",
note         = "Online",
}

@INPROCEEDINGS{ITD_COMSNETS24,
  author={Agarwal, Amit and Thombre, Anurag and Kedia, Kabir and Ghosh, Indrajit},
  booktitle={2024 16th International Conference on COMmunication Systems \& NETworkS (COMSNETS)}, 
  title={ITD: Indian Traffic Dataset for Intelligent Transportation Systems}, 
  year={2024},
  volume={},
  number={},
  pages={842-850},
  doi={10.1109/COMSNETS59351.2024.10427394}}

@INPROCEEDINGS{JATAYU_CONECCT24,
  author={Maji, Abhijnan and Preetham, Kummara and Ghosh, Indrajit},
  booktitle={2024 IEEE International Conference on Electronics, Computing and Communication Technologies (CONECCT)}, 
  title={JATAYU: A Large-scale Indian UAV Dataset for Vehicle Detection and Tracking}, 
  year={2024},
  volume={},
  number={},
  pages={1-6},
  doi={10.1109/CONECCT62155.2024.10677099}}

@article{Friedman2022TheVS,
  title={The Vendi Score: A Diversity Evaluation Metric for Machine Learning},
  author={Dan Friedman and Adji Bousso Dieng},
  journal={Trans. Mach. Learn. Res.},
  year={2022},
  volume={2023},
  url={https://api.semanticscholar.org/CorpusID:252715476}
}

@article{kappa1971fleiss,
author = {Fleiss, Joseph},
year = {1971},
month = {11},
pages = {378-382},
title = {Measuring nominal scale agreement among many raters},
volume = {76},
journal = {Psychological Bulletin},
doi = {10.1037/h0031619}
}

@article{observeragreement1977landis,
 ISSN = {0006341X, 15410420},
 URL = {http://www.jstor.org/stable/2529310},
 author = {J. Richard Landis and Gary G. Koch},
 journal = {Biometrics},
 number = {1},
 pages = {159--174},
 publisher = {International Biometric Society},
 title = {The Measurement of Observer Agreement for Categorical Data},
 urldate = {2026-04-06},
 volume = {33},
 year = {1977}
}
